\documentclass[runningheads]{llncs}

\usepackage{eccv}

\usepackage{eccvabbrv}

\usepackage{graphicx}
\usepackage{booktabs}
\usepackage{amssymb}
\usepackage{booktabs}
\usepackage{multirow}
\usepackage{wrapfig}

\usepackage[T1]{fontenc}
\usepackage{xcolor}
\usepackage{tcolorbox}
\tcbuselibrary{skins, listings}
\usepackage{listings}
\usepackage{caption}
\usepackage{float}

\usepackage[accsupp]{axessibility}  

\usepackage{hyperref}

\usepackage{orcidlink}

\begin{document}

\title{DeepGaze3.5-VL: Modeling Scanpaths via Autoregressive Token Prediction}

\titlerunning{DeepGaze3.5-VL}

\author{Susmit Agrawal\inst{1,2,3} \and
Matthias Bethge\inst{1,2} \and
Matthias Kümmerer\inst{1,2}}

\authorrunning{S.~Agrawal et al.}

\institute{$^1$University of T\"{u}bingen, $^2$T\"{u}bingen AI Center, $^3$IMPRS-IS}

\maketitle

\begin{abstract}
  Understanding human visual attention on a scene over time has applications in domains such as autonomous driving, interface design, inferring cognitive states. Modeling human visual scanpaths has historically relied on specialized architectures with hand-crafted geometric priors. While these architectures successfully model fixation sequences, their rigid structural biases restrict easy extendability and flexible conditioning. For instance, integrating task-specific instructions or adapting to distinct viewer identities traditionally requires custom, disjoint architectural additions. We address this limitation by framing scanpath prediction purely as a discrete sequence modeling task. By mapping spatial coordinates into an autoregressive vocabulary, we leverage the pretrained representations of Large Vision-Language Models (LVLMs). This formulation naturally absorbs diverse factors of variation: simple prompting modifications allow for global conditioning, such as providing viewer identities to capture personalized biases, or task-specific objectives (like free-viewing versus visual search). The framework can also easily integrate the modeling of per-fixation attributes, such as individual fixation durations alongside spatial locations. Crucially, this autoregressive alignment enables the scalable, exact computation of per-fixation log-likelihoods, directly equivalent to the commonly used Information Gain (IG) metric. Our model, \textbf{DeepGaze3.5-VL}, establishes a new state-of-the-art across multiple datasets, achieving 2.18 bits of IG on MIT1003, a 46\% improvement over matched-backbone DeepGaze III. This advantage persists even when baselines use identical high-capacity vision encoders.
  Beyond predictive performance, our generative framework serves as a powerful computational tool for direct behavioral interventions, allowing for controlled in-silico simulations that would be experimentally difficult or impossible to conduct in vivo. We demonstrate this ability by performing controlled interventions on the durations of pre-saccadic fixations, recovering known oculomotor phenomena purely from data.
\end{abstract}

\section{Introduction}
\label{sec:intro}
\noindent  Predicting human visual attention is a foundational challenge in computer vision, extending from static saliency to the task of scanpath prediction: modeling the ordered sequence of fixations humans make when exploring a scene. While saliency maps provide a time-averaged distribution of ``importance'', scanpaths reveal the underlying cognitive process of exploration. Accurately modeling this temporal dimension is vital for applications ranging from autonomous systems that must prioritize environmental cues~\cite{palazzi2018predicting} to human-computer interaction designs that align with natural viewing behavior~\cite{duchowski2002breadth}.

\noindent The static saliency literature has established that deep pretrained semantic representations outperform early hand-crafted features for predicting distributions of fixation regions~\cite{itti1998model,huang2015salicon,jia2020eml,kummerer2015deepgaze,linardos2021deepgaze}. Yet scanpath models have only partially 
\begin{wrapfigure}[26]{r}{0.5\textwidth}
\vspace{-2em}
    \centering
    \includegraphics[width=\linewidth]{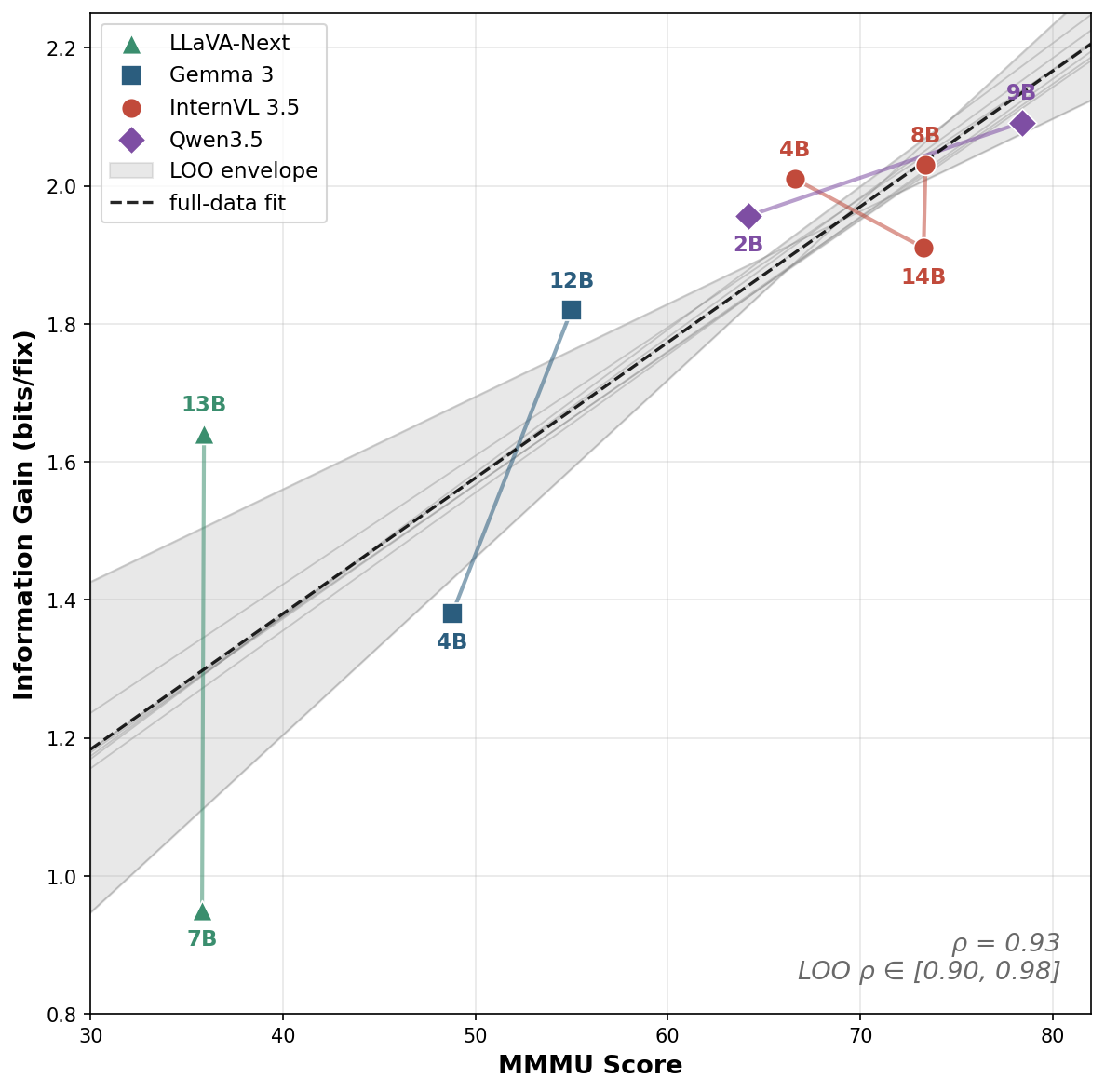}
    \caption{\textbf{Multimodal capability predicts scanpath quality.} Information Gain of LoRA-tuned models on MIT1003 versus MMMU score across four VLM families (LLaVA-Next, Gemma~3, InternVL~3.5, Qwen3.5) and multiple scales, with leave-one-out correlation envelopes. The strong correlation ($\rho = 0.93$) suggests that the representations underlying broad visual understanding are the same ones needed to predict human attention.}
    \label{fig:mmmu_vs_ig}
\end{wrapfigure}
absorbed this insight. Mechanistic models construct explicit dynamical systems for exploration (e.g., Constrained L\'evy Exploration~\cite{boccignone2004modelling}, saccadic biases~\cite{coutrot2016context}, object-based targeting~\cite{zelinsky2008theory}, inhibitory tagging~\cite{itti2000saliency}, or cognitive attention streams \cite{engbert2015scenewalk}); for a comprehensive review, see~\cite{kummerer2022state}. Existing end-to-end learning approaches such as DeepGaze~III~\cite{deepgaze3} invest heavily in architectural priors and augment static features with hand-crafted dynamic components that encode sequential dynamics. These approaches use task-specific architecture design rather than established sequential frameworks. More recent work has replaced these hand-crafted components with principled generative frameworks: diffusion models~\cite{cartella2025scandiff} and neural point processes~\cite{wang2025tppgaze} bring modern sequential modeling to the problem and consequently boost prediction performance, yet still train their generative machinery from scratch on gaze data alone. Across both generations of models, the pattern is the same: pretrained priors for space, trained-from-scratch machinery for temporal ordering.

\noindent We propose closing this gap with Large Vision-Language Models (LVLMs), which already provide sequential priors trained over billions of tokens of visually grounded reasoning. Drawing inspiration from recent VLM successes in geometric tasks~\cite{chen2021pix2seq,chen2023shikra,peng2023kosmos}, we reframe scanpath prediction natively as autoregressive sequence generation, predicting gaze as discrete coordinate tokens. This discretization allows the model to leverage its jointly learned semantic and sequential priors to model human fixation patterns. Our results indicate that scanpath prediction quality correlates strongly with general multimodal reasoning ability: across four VLM families and nine model scales, Information Gain tracks MMMU benchmark score with Spearman correlation $\rho = 0.93$ (minimum LOO $\rho = 0.90$, Figure~\ref{fig:mmmu_vs_ig}). This empirical trend strongly aligns with cognitive science theories positing that high-level semantic understanding and top-down task goals are the primary drivers of visual exploration~\cite{wolfe2021guided,zelinsky2008theory}. Our formulation also handles factors of variation, such as individual gaze patterns or task details natively using natural-language prefixes, unlike specialized architectures that require bespoke modules for task-conditioning~\cite{yang2022predicting} or individualized observer biases~\cite{Xue_2025,Jiang_2024}.

\noindent The autoregressive formulation also aligns scanpath prediction with the native training objective of foundation models, solving a major evaluation bottleneck. While recent continuous generative models~\cite{cartella2025scandiff,wang2025tppgaze} capture the stochastic variability of human gaze, they cannot natively compute exact spatial probability densities, forcing a reliance on heuristic sequence-alignment metrics like ScanMatch or MultiMatch. In contrast, we show that Information Gain ($IG$) is a direct manifestation of the next-token-prediction objective. This means that by optimizing next-token prediction, the model directly maximizes the exact information captured about human gaze behavior without surrogates or heuristics. We introduce an efficient metric computation strategy that decomposes the spatial joint probability to compute exact metric values from VLM outputs. Our fine-tuned models achieve up to 2.18 bits of IG on MIT1003, outperforming legacy specialists by up to 46\%. Our primary contributions are:

\begin{itemize}
\item \textbf{Formulating Scanpath Generation as Autoregressive Token Prediction:} By fine-tuning LVLMs to predict discrete coordinate tokens, we leverage pre-trained semantic priors to achieve a state-of-the-art 2.18 bits of Information Gain, outperforming the previous state-of-the-art with specialized geometric architectures by 46\%.

\item \textbf{Exact and Efficient Evaluation for Generative Scanpath Models:} We exploit the autoregressive structure of the VLM to decompose the full spatial joint probability as a product of marginals over coordinate tokens. We show that this formulation enables rapid and mathematically exact Information Gain computation at scale.

\item \textbf{Cross-Domain Generalization, Flexible Conditioning, and Interventional Analysis:} We demonstrate that DeepGaze3.5-VL generalizes out-of-distribution across disparate datasets and successfully adapts to novel datasets in few-shot. We further show that the model can be easily conditioned on different factors of variation. Additionally, the formulation naturally allows for counterfactual interventions entirely in-silico, enabling analysis of human gaze patterns without additional data collection.

\end{itemize}

\section{Methodology}
\label{sec:methods}
\label{subsec:autoregressive_formulation}

We formulate scanpath prediction as conditional sequence generation. Given an image $I \in \mathbb{R}^{H \times W \times 3}$, we aim to predict the human scanpath $\mathcal{S} = \{(x_1, y_1), (x_2, y_2), \\ \ldots, (x_n, y_n)\}$, where each $(x_i, y_i)$ represents the spatial coordinates of the $i$-th fixation in the viewing sequence. We implement this using Vision-Language Models (VLMs) parameter-efficiently fine-tuned via Low-Rank Adaptation (LoRA)~\cite{lora}; we refer to our resulting model as DeepGaze3.5-VL. The VLM encodes the image through a pretrained vision encoder and generates coordinate sequences as structured text outputs alongside optional conditioning factors $\mathcal{C}$ (e.g., observer biases or task descriptions) natively in the prompt.

\noindent A critical challenge in using language models for coordinate prediction is ensuring fair probabilistic comparison. Different coordinate values may tokenize to different numbers of tokens depending on their magnitude (e.g., ``7'' versus ``73''), confounding likelihood computations independent of actual spatial probability. We address this globally by encoding coordinates as zero-padded two-digit integers ranging from 00 to 99, representing percentage positions along each image dimension. We verified empirically that in the InternVL3.5 tokenizer, every valid coordinate digit is represented identically (e.g., ``05'' $\rightarrow$ [``0'', ``5'']). This guarantees that each fixation contributes exactly four coordinate tokens to the sequence, ensuring uniform coordinate length regardless of spatial position.

\noindent With uniform tokenization guaranteed, the model is trained via standard next-token prediction on ground-truth scanpaths. Because the VLM predicts logits over its token vocabulary, this factorization provides the exact autoregressive log-likelihood of any valid fixation coordinate sequence as:
\begin{align}
\log P(\mathcal{S} \mid I, \mathcal{C}) &= \sum_{i=1}^{n} \log P\big((x_i, y_i) \mid I, \mathcal{C}, (x_1, y_1), \ldots, (x_{i-1}, y_{i-1})\big)
\label{eq:autoregressive}
\end{align}
This property enables rigorous information-theoretic computation: measured differences in likelihood directly reflect the model's learned visual attention patterns rather than arbitrary tokenization artifacts.

\subsection{Evaluation Metrics}
\label{subsec:metrics}

\paragraph{Information Gain.} Information Gain measures improvement in Log-Likelihoods of model predictions over a baseline model (usually the centerbias, as described by~\cite{tatler2007centerbias,kummerer2015information}) in bits per fixation, computed as $\text{IG} = (\text{LL}_{\text{model}} - \text{LL}_{\text{centerbias}}) / \ln(2)$. Each dataset has its own independent centerbias baseline, which is held fixed and applied uniformly across all models when computing IG. A positive IG indicates the model captures information about attention beyond this spatial prior. Information Gain is the standard metric for saliency model comparison~\cite{kummerer2015information} and enables meaningful comparison across datasets without any heuristics.

For a scanpath with $n$ fixations encoded as a token sequence, this is computed as $\text{LL} = \sum_{i=1}^{n} \sum_{j=1}^{4} \log P(t_{i,j} \mid I, \mathcal{C}, t_{< (i,j)})$, where $t_{i,j}$ is the $j$-th token of the $i$-th fixation coordinate (four tokens per fixation: two for $x$, two for $y$). Because the VLM's full vocabulary extends far beyond the ten digit tokens used for coordinate encoding, raw logits leak mass to non-digit tokens. We eliminate this leakage by restricting and renormalizing: at every digit position, we extract logits of each valid digit and apply a log-sum-exp correction over the digit set, $\log \hat{P}(t_{i,j}) = \log P(t_{i,j}) - \log \sum_{d \in \{0,\ldots,9\}} P(d)$, yielding exact spatial densities. This renormalization is applied identically in both the full prefix-tree evaluation and the rapid IG evaluation described below. 

\paragraph{AUC.} Area Under the RoC Curve (AUC) is a widely used spatial metric that measures ranking performance: whether fixated locations score higher than non-fixated locations in the predictive probability map. Unlike IG, AUC evaluates only the relative ordinal ranking of probabilities and ignores the calibration of the actual probability mass. Consequently, models can achieve high AUC scores despite being poorly calibrated or overly confident (e.g., assigning excessively sharp probability peaks to only top locations), particularly when the spatial densities have multiple modes competing for visual attention. While we report AUC for completeness and comparison with prior ranking-optimized models, we emphasize IG as the primary metric because it rewards well-calibrated probabilistic predictions that reflect genuine uncertainty in human visual behavior.

\subsection{Model Evaluation at Scale}

\begin{figure}[t]
    \centering
    \includegraphics[width=\linewidth]{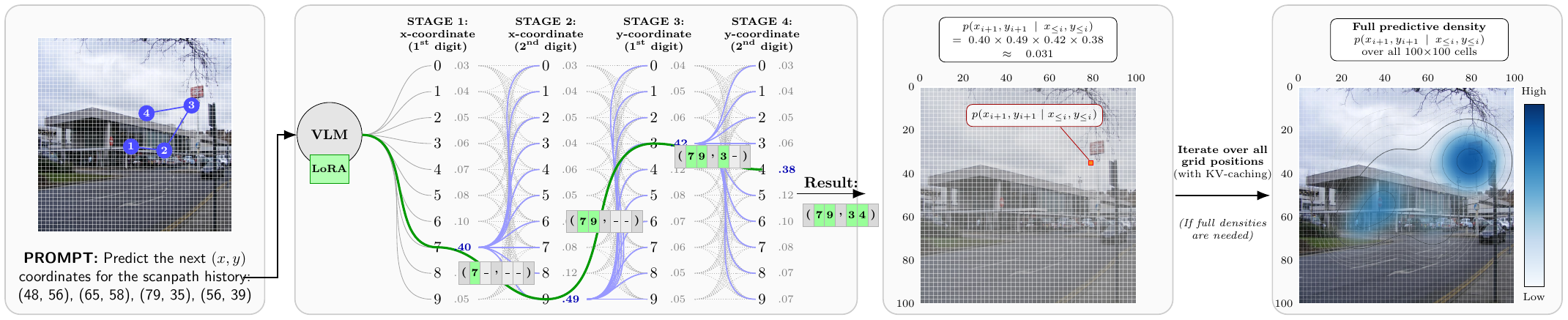}
    \caption{\textbf{Schematic Diagram of our Inference Pipeline.} The diagram illustrates how our VLM-based framework computes the likelihood of specific spatial coordinates for evaluation metrics like Information Gain, alongside the procedure for sequentially sampling full scanpaths, where each scanpath corresponds to traversing a specific path through the autoregressive token tree. This unified formulation naturally accommodates both point-wise probability estimation and unconstrained generative exploration from the same underlying predictions.}
    \label{fig:schematic}
\end{figure}

To compute spatial metrics (such as AUC), we require the model's full spatial probability distribution $P(x, y)$ across the $100 \times 100$ coordinate grid. With $10{,}000$ unique coordinate combinations, a naive strategy based on teacher-forcing coordinate values one at a time requires $10{,}000$ full-context forward passes per fixation. For an 8B-parameter generative model, this is computationally prohibitive for standard eye-tracking datasets. We visualize both evaluation strategies and the scanpath sampling procedure in Figure~\ref{fig:schematic} (see Appendix~\ref{sec:app_implementation} for detailed implementation and timing results). 
 
\paragraph{Exact Evaluation via Prefix Tree.}
We exploit the autoregressive structure of our coordinate tokenization to avoid independent sequence scoring of Monte-Carlo approaches entirely. Using the chain rule, we decompose the joint spatial probability as $P(x, y) = P(x_1) \cdot P(x_2 \mid x_1) \cdot P(y_1 \mid x_1, x_2) \cdot P(y_2 \mid x_1, x_2, y_1)$. By caching Key-Value states at shared prefixes, we evaluate this as a 4-phase prefix tree, expanding only valid digit branches ($0$–$9$) at each stage. The $x_1$ marginal requires $1$ forward pass, the $x_2$ conditional requires $10$, the $y_1$ conditional requires $100$, and the $y_2$ conditional requires $1{,}000$, yielding the complete $10{,}000$-cell probability map in $1{,}111$ passes over the prefix without requiring approximations. We upscale the $10{,}000$-cell probability map to original image resolution and renormalize before metric computations (Figure~\ref{fig:schematic}, left).

\paragraph{Rapid IG Evaluation.}
For settings requiring only information-theoretic metrics, we can optimize evaluation to comprise exactly $4$ forward passes per fixation. IG depends only on the probability assigned to the ground-truth sequence; it does not require generating alternative spatial hypotheses. By conditioning the model on the ground-truth fixation-history at each step, we can extract the exact scalar $P(x_{\text{GT}}, y_{\text{GT}})$ and use it to compute IG at minimal cost. We empirically verify that the IG values obtained in this way match the full-grid IG values.

\section{Experiments and Results}
\label{sec:expt}
Our experiments use InternVL3.5-8B tuned with LoRA~\cite{lora}, and are designed to answer several questions to holistically evaluate VLM capabilities in modelling human scanpaths. First, can VLMs model scanpaths well enough to outperform state-of-the-art attention models? Second, is this due to better visual features, or do improvements come from elsewhere? Third, do VLMs learn the sequential structure of scanpaths, or merely improved spatial distributions? Fourth, what capabilities do they offer beyond existing state-of-the-art scanpath models? Fifth, what can they tell us about human visual attention?

\subsection{Experimental Setup}
\label{subsec:setup}

We evaluate our models on five diverse, publicly available eye-tracking datasets to ensure robustness across different image statistics and viewing behaviors: MIT1003~\cite{mit_saliency}, COCO-FreeView~\cite{chen2021coco_search18}, CAT2000~\cite{borji2015cat2000}, DAEMONS~\cite{david2024potsdam}, and FiGrIm~\cite{bylinskii2015intrinsic}. We provide additional dataset and training details in Appendix~\ref{sec:app_implementation}. 

\paragraph{Baselines.} The \textit{centerbias} predicts fixations according to the dataset-specific spatial prior, capturing the tendency to look near image centers. \textit{TPPGaze}~\cite{wang2025tppgaze} uses neural temporal point processes to jointly model the spatial and temporal dynamics of saccades. \textit{HAT}~\cite{yang2022predicting} is a transformer-based architecture for task-driven and free-viewing attention prediction. \textit{DeepGaze III}~\cite{deepgaze3} extends the DeepGaze series of spatial models to predict scanpaths, using separate branches for encoding spatial densities and fixation histories. 
While we discussed other scanpath models in previous sections, most of them do not produce saliency maps and cannot be compared based on information-theoretic metrics.

\noindent We evaluate three experimental configurations: \textit{combined}, which jointly trains on training splits of all datasets mixed; \textit{single-dataset}, which trains and tests on each dataset individually; and \textit{leave-one-dataset-out} (LODO), which trains on four datasets and tests on the held-out set of the fifth to measure out-of-distribution generalization. All models and baselines are evaluated on the same held-out splits across datasets.

\subsection{Main Results}
\label{sec:main_results}

Table~\ref{tab:main_comparison_results} presents our main results evaluated with Information Gain (IG) and Area Under the Curve (AUC). Our combined training approach yields substantial state-of-the-art performance, achieving 2.18 bits of IG on the MIT1003 dataset---a 46\% improvement over the matched-backbone DeepGaze~III baseline (1.49 bits), with parallel results across all other datasets. We also visualize sampled scanpaths from different models in Appendix~\ref{sec:app_qualitative}, showing that DeepGaze3.5-VL produces scanpaths that are qualitatively more similar to human scanpaths.

\begin{table}[t]
\centering
\caption{\textbf{Main comparison across five eye-tracking datasets.} Information Gain (IG, bits/fixation) and AUC for baselines and DeepGaze3.5-VL under in-distribution (IID-Only), combined multi-dataset (Combined), and leave-one-dataset-out (OOD) training. $^{*}$~denotes models \emph{not} trained on the target dataset. DeepGaze~III--6B is retrained with the InternViT-6B backbone for a fair feature-matched comparison.}
\label{tab:main_comparison_results}
\scriptsize
\begin{tabular}{c|cc|cc|cc|cc|cc}
\toprule
\textbf{Method} & \multicolumn{2}{c|}{\textbf{MIT1003}} & \multicolumn{2}{c|}{\textbf{COCO-FV}} & \multicolumn{2}{c|}{\textbf{CAT}} & \multicolumn{2}{c|}{\textbf{Daemons}} & \multicolumn{2}{c}{\textbf{Figrim}} \\
 & IG & AUC & IG & AUC & IG & AUC & IG & AUC & IG & AUC \\
\midrule
Center Bias & 0.00 & 0.81 & 0.00 & 0.81 & 0.00 & 0.85 & 0.00 & 0.84 & 0.00 & 0.80 \\
TPPGaze~\cite{wang2025tppgaze} & 1.18 & 0.90 & 1.41 & 0.92 & - & - & 1.52$^{*}$ & 0.88$^{*}$ & - & - \\
HAT & 1.25$^{*}$ & 0.92$^{*}$ & 1.39 & 0.93 & 1.07$^{*}$ & 0.92$^{*}$ & 1.53$^{*}$ & 0.89$^{*}$ & 0.99$^{*}$ & 0.91$^{*}$ \\
DeepGaze III - 6B & 1.49 & 0.91 & 1.69 & 0.92 & 1.60 & 0.93 & 2.32 & 0.92 & 1.59 & 0.92 \\
\midrule
\textbf{DeepGaze3.5-VL (LODO)} & 1.94$^{*}$ & \textbf{0.94}$^{*}$ & 1.79$^{*}$ & 0.93$^{*}$ & 1.66$^{*}$ & 0.94$^{*}$ & 2.30$^{*}$ & 0.93$^{*}$ & 1.73$^{*}$ & 0.93$^{*}$  \\
\textbf{DeepGaze3.5-VL (IID-Only)} & 2.03 & \textbf{0.94} & 2.15 & \textbf{0.94} & 2.01 & \textbf{0.95} & 2.67 & \textbf{0.94} & 1.88 & 0.93 \\
\textbf{DeepGaze3.5-VL (Combined)} & \textbf{2.18} & \textbf{0.94} & \textbf{2.19} & \textbf{0.94} & \textbf{2.07} & \textbf{0.95} & \textbf{2.75} & \textbf{0.94} & \textbf{1.93} & \textbf{0.94} \\
\bottomrule
\end{tabular}
\vspace{-10pt}
\end{table}

\paragraph{The Matched-Backbone Control.} To test if VLM success is solely due to better visual features from larger, more modern vision encoders, we retrain DeepGaze III with the InternViT-6B encoder used by InternVL3.5-38B. Note that this is a larger and more capable vision encoder than the one used in our VLM (InternViT-300M). If visual features explained the improvement, matched backbone DeepGaze III should approach VLM performance.
\begin{wrapfigure}{L}{0.5\textwidth}
\vspace{-3em}
    \centering
    \includegraphics[width=\linewidth]{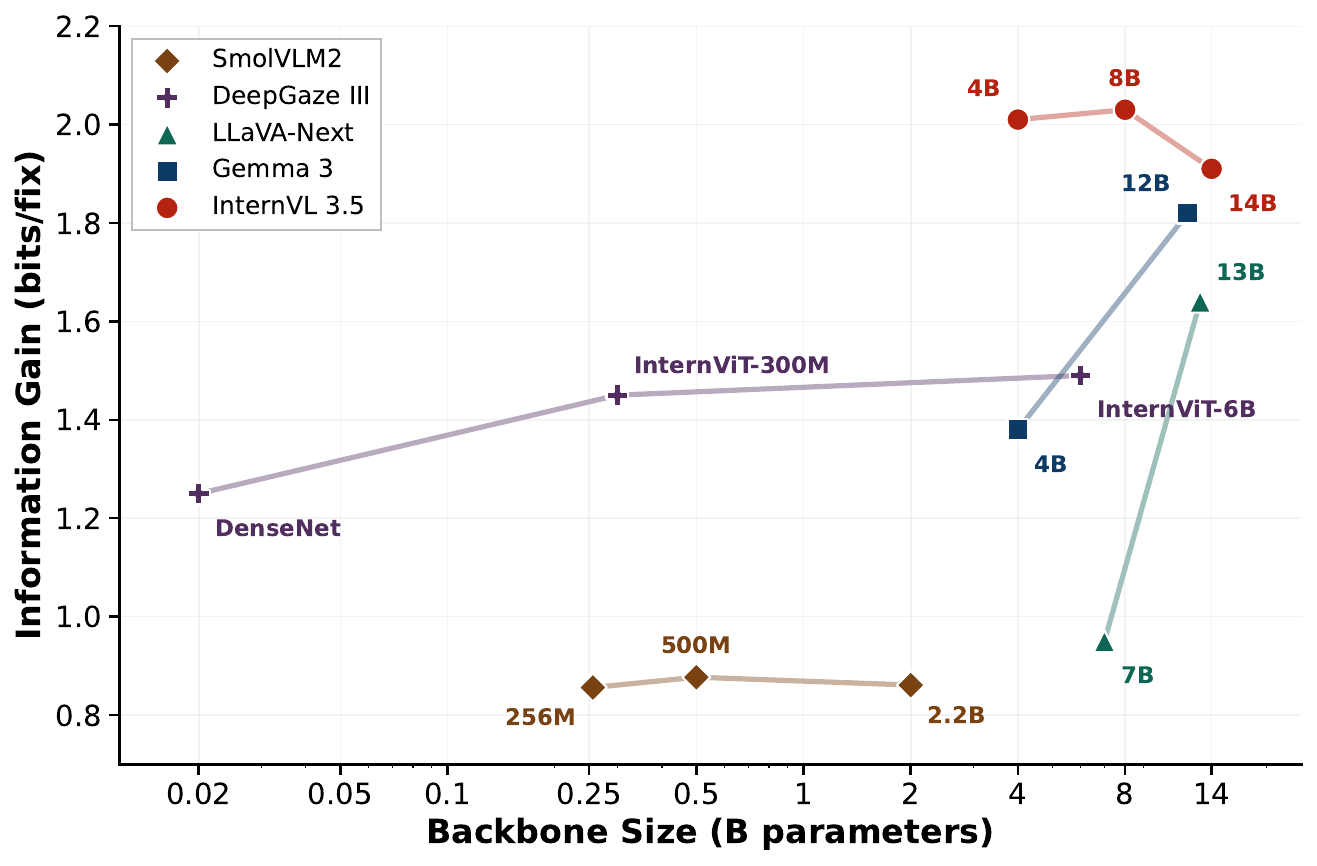}
    \vspace{-2em}
    \caption{\textbf{Architecture matters more than raw scale.} SmolVLM2 models remain flat regardless of scale; DeepGaze~III improves with better vision encoders; other VLM families show steep gains with size but with markedly different efficiency. InternVL~3.5 reaches 2~bits at 4B parameters, while LLaVA-Next requires 13B for 1.6~bits.}
    \label{fig:size_vs_ig}
\vspace{-4em}
\end{wrapfigure}
Our results establish that this is not the case, and that there is a significant gap between the contributions of the isolated vision encoder and the complete VLM model. For instance, matched-backbone DeepGaze III achieves 1.49~bits IG on MIT, substantially below InternVL3.5-8B despite using better visual representations. This is also consistent across other datasets. For completeness, we also test the InternViT-300M-backed DeepGaze III on MIT, getting an IG of~1.45 bits. The improvement, therefore, cannot be attributed to visual features alone. The language model of the VLM captures additional information that DeepGaze's saliency-focused training cannot capture. This also reflects in the performance of the two LLaVA model variants in Figure~\ref{fig:mmmu_vs_ig}, which use identical vision encoders and only differ in their LLM branch sizes. Figure~\ref{fig:size_vs_ig} extends this analysis across model families and scales. Raw size does not explain gains: SmolVLM2 models have poor performance regardless of scale (0.85~bits from 256M to 2.2B weights), while InternVL~3.5-4B already achieves 2~bits, matching or exceeding models $3\times$ its size. This contrasts with findings in static saliency, where scaling backbone size yields diminishing returns~\cite{khanuja2025modeling}; in our autoregressive setting, the quality of learned multimodal representations, as evidenced by the strong $\rho = 0.93$ correlation (minimum LOO $\rho = 0.90$) between MMMU score and IG (Figure~\ref{fig:mmmu_vs_ig}), remains a significant driver of scanpath prediction performance. 


\paragraph{Out-of-Domain Generalization.} The ``DeepGaze3.5-VL (LODO)'' row in Table~\ref{tab:main_comparison_results} presents LODO results, where models trained on four datasets are tested on the fifth. Models maintain substantial performance on held-out datasets, suggesting transferable visual attention patterns rather than dataset-specific biases. VLMs match or outperform DeepGaze~III-6B in-distribution performance, despite never having seen the target dataset during training. Zero-shot OOD performance can be further optimized with minimal data, as detailed in Appendix~\ref{sec:app_fewshot}.

\subsection{Isolating Spatial and \\ Sequential Contributions}
\label{subsec:sequential}

\begin{wrapfigure}[14]{r}{0.5\textwidth}
    \vspace{-7em}
    \centering
    \includegraphics[width=\linewidth]{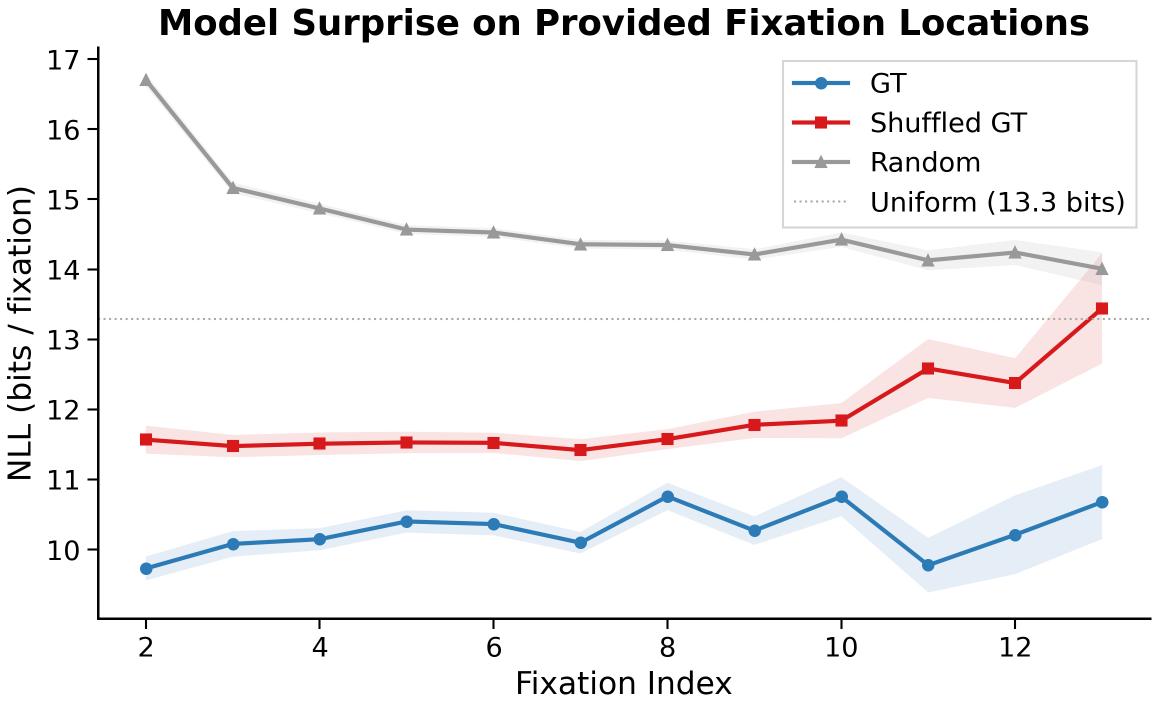}
    \vspace{-2em}
    \caption{\textbf{Per-fixation surprise over time.} Ground-truth surprise (NLL, bits) remains stable; surprise for shuffled sequences begins near ground truth but diverges later: the model is sensitive to sequential dependencies. Random sequences sit above the uniform baseline, as the model concentrates mass on scene-relevant regions, inflating surprise for uniformly sampled fixations.}
    \label{fig:nll_curves}
\end{wrapfigure}

We now analyze how much of the model's predictive advantage stems from knowing \emph{where} observers look, versus knowing historic saccade dynamics. To isolate the gain from ordering rather than better spatial saliency, we conduct the following experiment.
For each human scanpath in the MIT held-out set, we take three variants. (1) \textbf{Ground truth (GT)}: the human fixation sequence. (2) \textbf{Shuffled GT}: the same fixations in uniformly random order. (3) \textbf{Random}: coordinates sampled uniformly from the $100 \times 100$ grid, matching the original sequence length. 


\noindent The per-fixation dynamics (Figure~\ref{fig:nll_curves}) illustrate the importance of sequential dependencies. Ground-truth surprise (Negative Log-Likelihood) is remarkably stable throughout the scanpath. Note that the surprise is monotonic with \textit{perplexity}\cite{perplexity} used for quantifying LLM uncertainty. Shuffled GT surprise begins close to ground truth: early fixations are well-predicted by static spatial salience. However, it diverges significantly for later fixations; the accumulated violations of temporal dependencies become increasingly apparent to the model, proving that late-stage fixations cannot be explained merely as draws from a static spatial density. Random scanpath surprise for the model sit well above the surprise of even a uniform predictor. This is expected: because the model concentrates probability mass on scene-relevant regions, uniformly random fixations usually land in the model's low-probability areas, inflating measured surprise.

\begin{wrapfigure}{l}{0.5\textwidth}
    \vspace{-2em}
    \centering
    \includegraphics[width=\linewidth]{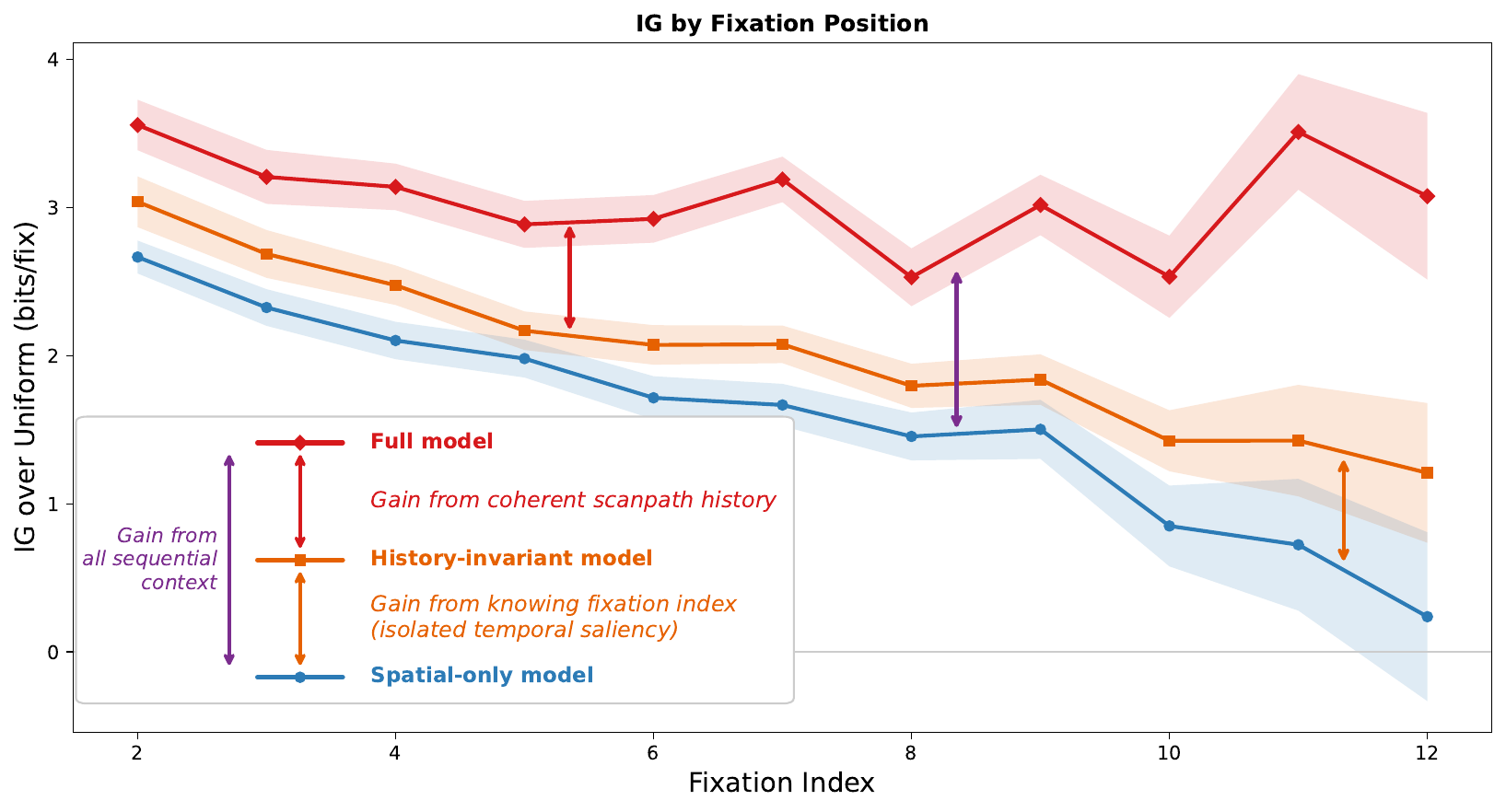}
    \vspace{-2em}
    \caption{\textbf{Evolution of spatial and sequential information across scanpaths.} To isolate contributions of spatial and sequential knowledge, we compare three models. The \textbf{Spatial-only model} decays to near-zero predictability over time, confirming late-stage viewing requires ordinal knowledge. The \textbf{History-invariant model} learns population-level ordinal tendencies, yielding a stable saliency advantage. The \textbf{Full model} isolates the value of idiosyncratic scanpath coherence, dominating late-stage fixation predictability.}
    \label{fig:spatiotemporal_ig}
    \vspace{-2em}
\end{wrapfigure}

\noindent While Figure~\ref{fig:nll_curves} establishes that order matters overall, it masks a highly dynamic process. To understand \emph{how} spatial and sequential information evolve across a scanpath, we compare ablated models explicitly data-constrained during training. Figure~\ref{fig:spatiotemporal_ig} decomposes the informative features of scanpaths using three distinct models: \textbf{(1) The Spatial Foundation (Spatial-only model):} This model is trained to predict fixations with points sampled from the entire fixation pool across all subjects for the given image as history. This means that the model cannot learn to use any sequential priors for predicting the next fixation. While it performs well initially, its predictive performance decays rapidly from $\sim$2.7 bits to $\sim$0.25 bits by fixation 12. Image content is the primary driver of early orienting, but rapidly loses predictive power as primary targets are exhausted.
\textbf{(2) History Invariant Ordinal Saliency:} Training scanpath samples for the History-invariant model are created by independently sampling each $i$-th fixation from an aggregated pool of all $i$-th human fixations for the input image. This encodes population-level tendencies for \emph{when} regions correlate with a certain fixation index. The stable $\sim$0.3--0.6 bit gap over the spatial-only baseline shows that even without knowing what an individual observer has examined, knowing \emph{how long} they have been exploring the scene provides a consistent predictive signal. This is consistent with prior findings that knowledge of temporal bins of fixations helps predict better saliency, with the first temporal slice being the most informative \cite{aydemir2023tempsal} and encoding primarily spatial content \cite{heiko2019saliency}. \textbf{(3) Idiosyncratic Scanpath Coherence:} The Full model sees the true history of the current observer. The growing performance gap between the Full and History-invariant models isolates the value of statefulness--knowing the past observer choices becomes a critical predictor of their next move, reflecting explicit mechanistic constraints proposed in cognitive literature, such as Inhibition of Return~\cite{itti2000saliency,engbert2015scenewalk} and systematic coverage strategies or foraging~\cite{boccignone2004modelling,coutrot2016context}. Interestingly, although spatial contribution drops massively in late fixations, overall predictability stays mostly constant given viewing history, maybe even with an increase towards the end(fixations 9--11). Deep into a scanpath, the visual stimulus itself matters less than the observer's internal, unfolding state, a dynamic that the autoregressive framework seamlessly captures.


\subsection{Duration-Conditioned Prediction}
\label{sec:duration_conditioning}

A key temporal feature of human scanpaths is fixation duration - the amount of time the eye remains stationary before initiating the next saccade. To
\begin{wrapfigure}[18]{r}{0.5\textwidth}
    \vspace{-2em}
    \centering
    \includegraphics[width=\linewidth]{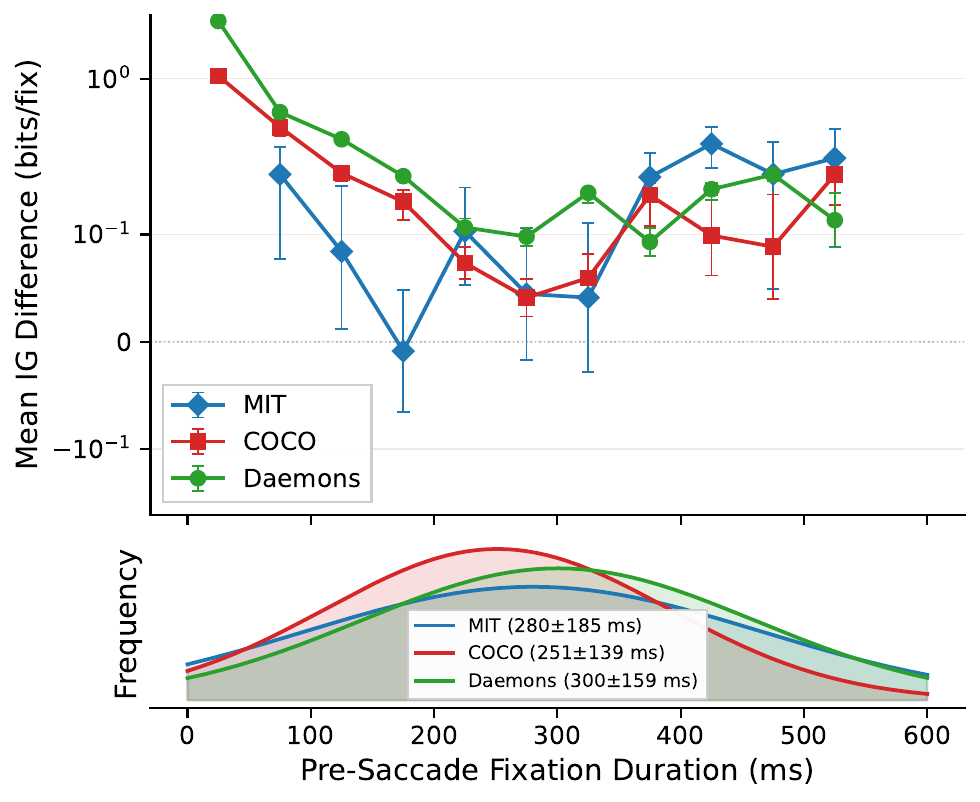}
    \vspace{-2em}
    \caption{\textbf{Duration helps predict early fixations.} The value of conditioning on duration is asymmetric, yielding the highest gains for short fixations, and some improvement in predictability of very late fixations.}
    \label{fig:duration_contrib}
\end{wrapfigure}
investigate whether VLMs can leverage this temporal signal to improve spatial predictions, we introduce duration conditioning. We adapt our tokenization scheme by augmenting the spatial coordinate format with discretized duration values. We quantize fixation durations into 1ms bins, appended to the prediction alongside the coordinate tuples ($(x, y) \rightarrow (x, y, d)$). We train these models on MIT, COCO-FV, and Daemons and evaluate their spatial prediction performance given the durations and spatial locations of previous fixations. We compare against the duration-agnostic models from Sec.~\ref{sec:main_results}, showing that duration conditioning improves spatial IG across all three datasets: from 2.03 to 2.12 bits on MIT1003, from 2.15 to 2.30 bits on COCO-FV, and from 2.67 to 2.84 bits on Daemons.
\vspace{-1em}

\begin{figure*}[t]
  \centering
\includegraphics[width=\linewidth]{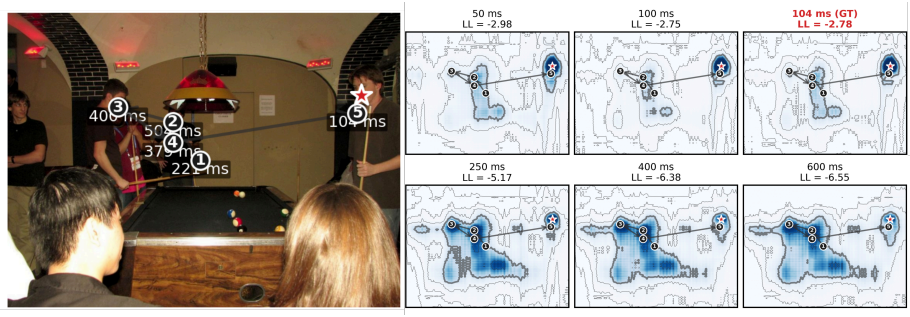}
  \caption{%
    \textbf{Counterfactual duration intervention.}
    Single fixation transition with the conditioned
    source-fixation duration swept from 50\,ms to 600\,ms; the
    ground-truth duration is marked in \textcolor{red}{red}. All other
    inputs (image, spatial history, model weights) are identical across
    panels. The model concentrates mass sharply at
    50--100\,ms and broadens monotonically beyond 250\,ms.%
  }
  \label{fig:density_sweep}
  \vspace{-2em}
\end{figure*}

\paragraph{Counterfactual duration intervention.} A key feature of our framework is allowing counterfactual interventions \emph{in-silico}, which are difficult or impossible to perform \emph{in-vivo}. We demonstrate this by evaluating a single fixation transition while holding the image, spatial history, and model weights fixed, and sweeping only the duration of the most recent fixation from 50\,ms to 600\,ms. 

\noindent The results, shown in Fig.~\ref{fig:density_sweep}, demonstrate a structured, duration-dependent modulation of the spatial distribution. At short conditioned durations (50--100\,ms), the model produces predictions close to the preceding fixation, with some salient predictions aimed at distant regions, such as the moderate-density region near the image center. This behavior corresponds to the \emph{ambient} processing mode~\cite{unema2005,velichkovsky2005}. As the conditioned duration increases beyond 250\,ms, the overall spatial entropy 
increases from 9.56\,bits at 50\,ms to 10.68\,bits at 600\,ms, with a pronounced step between 250 and 400\,ms (+0.68\,bits). However, despite this broadening, a high-density prediction remains close to the immediately preceding fixation, corresponding to the \emph{focal} processing mode~\cite{unema2005,velichkovsky2005}. The duration-agnostic model's entropy (9.80\,bits) coincides predictably with the 250-300ms region, reflecting an implicit aggregate model over median durations.

\noindent This asymmetric value of duration information (Fig.~\ref{fig:duration_contrib}) exhibits additional emergent functional alignment with known oculomotor phenomena. Short fixations ($<$100\,ms) yield larger IG as they select for a low-entropy behavioral regime, associated with pre-programmed saccades (parallel programming~\cite{becker1979} or saccadic momentum~\cite{smith2009,coutrot2016context}). Conversely, long durations show lower IG boosts compared to shorter durations as they indicate a high-entropy regime of deliberated saccade planning. 

\noindent The practical consequence is significant: the VLM natively recovers the behavioral signatures purely from the statistical regularities of gaze data, functioning as an \emph{in-silico} sandbox for testing how variables modulate spatial outcomes. We emphasize that while modeling duration is a compelling demonstration, this capability naturally extends to testing any factor of interest.

\noindent 
\begin{wrapfigure}{l}{0.5\textwidth}
    \vspace{-2em}
    \centering
    \includegraphics[width=\linewidth]{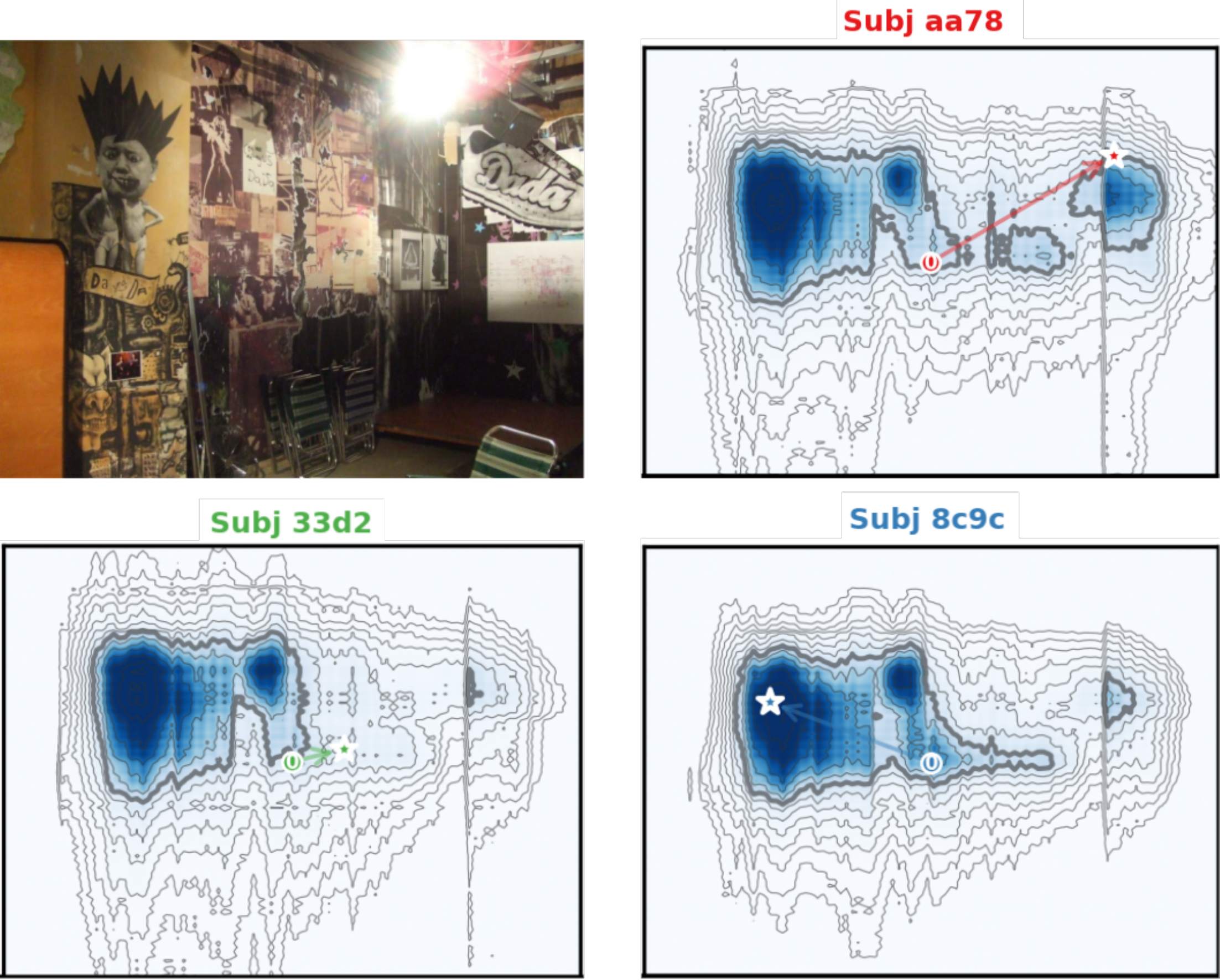}
    \caption{\textbf{Subject IDs modulate spatial distributions.} Predicted density over the first post-central fixation for three subject identifiers on an image from MIT1003. Differences are purely due to the subject token.}
    \label{fig:subject_first_fixation}
    \vspace{-2em}
\end{wrapfigure}

 \vspace{-2em}
\subsection{Subject Conditioning}
\label{sec:subject_conditioning}

A key advantage of our formulation is the ease with which additional conditioning variables can be incorporated. To demonstrate this, we extend our framework to model individual differences in viewing behavior \cite{benjamin2019subject} by conditioning on subject identity, requiring no architectural modifications whatsoever. During training, each scanpath prompt is augmented with an anonymized subject identifier, e.g., \textit{``Predict the scanpath by subject a3f8 for the image.''}, and randomly masking 10\% of the training samples with a generic \texttt{<unknown>} token. At test time, providing the true subject token yields a subject-specific prediction, while using the \texttt{<unknown>} token produces a subject-agnostic prediction from the same model. We evaluate on the MIT1003 dataset using 5-fold leave-subject-out cross-validation. Note that this requires different training and held-out splits from the previous experiments, as we also evaluate the model's capabilities to predict scanpaths of \emph{unknown} individuals.

\noindent Providing the subject identity improves IG by $+0.09$~bits/fixation on average. The effect is modest in absolute terms, approximately 5\% of total Information Gain, but consistent with the well-established finding that image content accounts for the majority of variance in fixation behavior while individual differences contribute a smaller but systematic component~\cite{dehaas2019}. 

\noindent As shown in Figure~\ref{fig:subject_first_fixation}, the model learns distinct spatial priors per observer from the very first fixation. The magnitude of this inter-subject divergence is strongly image-dependent: 
it emerges most strongly when scenes afford multiple plausible viewing strategies rather than a single dominant salient object. Detailed analysis of how these priors evolve and diverge over successive fixations is provided in Appendix~\ref{sec:app_subjectwise}.
\vspace{-1em}
\subsection{FreeView vs.\ Visual Search}
\label{sec:search}
\vspace{-0.5em}
The text-based conditioning approach naturally extends to different viewing tasks. To demonstrate this, we train a search-conditioned model on COCO-Search18~\cite{chen2021coco_search18}, which pairs COCO images with goal-directed scanpaths for 18 object categories. The model receives the search target in its text prompt (e.g., ``\texttt{...predict a scanpath searching for a \textit{clock}...}'') and is otherwise identical to the free-viewing model in architecture and training procedure.

\begin{wrapfigure}{r}{0.5\textwidth}
    \vspace{-2em}
    \centering
    \includegraphics[width=\linewidth]{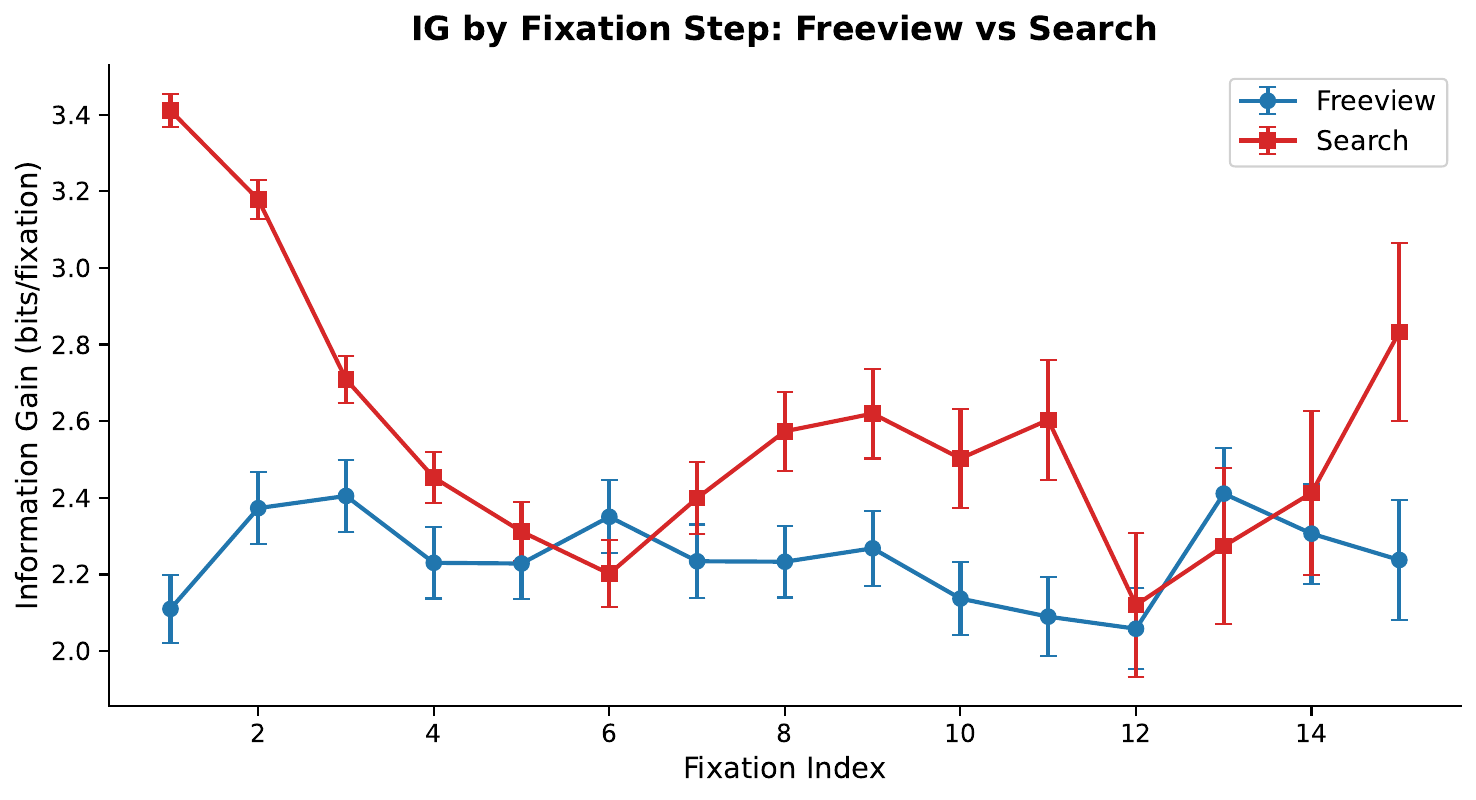}
    \vspace{-2em}
    \caption{\textbf{Per-fixation IG in Task vs Free-Viewing.} Search IG is high initially: viewing patterns are established early in search.}
    \label{fig:per_fixation_ig}
    \vspace{-2em}
\end{wrapfigure}

\noindent The search-conditioned model achieves an aggregate IG of $3.27$~bits/fix, substantially exceeding the free-viewing model's $2.15$~bits/fix on the same COCO images. The task effect is almost entirely mediated by target presence. When the search target appears in the image, IG rises to $4.23$ bits/fix; when it is absent, IG drops to $2.29$ bits/fix, which is close to free viewing predictability. For comparison, HAT~\cite{yang2022predicting}, the strongest existing search-conditioned baseline, achieves $3.03$ bits/fix on target-present and $1.68$ bits/fix on target-absent trials, placing it well below DeepGaze3.5-VL in both conditions. 

\noindent Our results show that on target-absent trials, observers fall back to exhaustive scanning that resembles free exploration, and the search prompt provides minimal additional constraint. As detailed in Appendix~\ref{sec:app_search}, this pattern holds across all individual target categories: target-present IG vastly outperforms target-absent IG, confirming that the predictive advantage relies heavily on visual guidance to a successful find (see Figure~\ref{fig:target_ig}). Notably, this predictive power generalizes in a zero-shot manner: unseen target categories achieve comparable IG to seen targets, demonstrating the VLM's ability to seamlessly leverage open-vocabulary semantic knowledge without object-specific retraining.

\begin{figure}[ht]
    \centering
    \includegraphics[width=\linewidth]{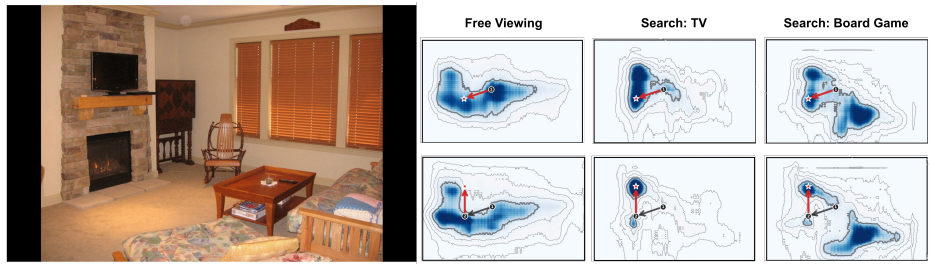}
    \caption{\textbf{Task changes predicted density without architectural changes.} Left: a COCO living-room scene. First col: free-viewing density maps at fixations 1 \& 2 spread probability broadly across the scene. Second col: visual search (target: TV) concentrates density near the target from fixation~1 onward. Last col: visual search (target: Board Game) extends to unseen targets in zero-shot. Target: TV scanpath fixations used as history for density predictions in all three cases.}
    \label{fig:fv_vs_search}
\end{figure}

\noindent Per-fixation analysis (Figure~\ref{fig:per_fixation_ig}) reveals that search IG is elevated from the first saccade onward, with no initial ramp-up period. The model captures immediate target-directed orienting rather than a gradual transition from scene exploration. Free-viewing IG is lower and flatter, consistent with the broader, less constrained nature of exploratory gaze. Both trajectories align with known phenomena in visual search~\cite{wolfe2021guided}: early saccades in search are target-informed, and target-absent trials produce quasi-random patterns~\cite{zelinsky2008theory} similar to free-viewing. 

\noindent Qualitatively (Figure~\ref{fig:fv_vs_search}), the free-viewing model spreads probability broadly across the scene, while the search model concentrates density near the target from fixation~1 onward. The model recovers these task-dependent dynamics purely from data, without any search-specific modules.

\section{Discussion}
\label{sec:discussion}
\paragraph{Implications for Cognitive Modeling.} Our framework enables using generative models as \emph{in-silico} experimental platforms for cognitive science, potentially opening new research avenues. As demonstrated by our duration intervention (Sec.~\ref{sec:duration_conditioning}), researchers can systematically vary cognitive variables and observe the spatial response. Durations in free-viewing experiments are involuntary, and forcing a fixation duration (if possible at all) will change viewing behavior, as observers will not be "free-viewing" anymore. Such experimental studies usually need to use extreme caution to avoid correlations and confounders, and it's likely that some confounding factors nevertheless leak into behavior. Simulation-based counterfactual experimentation has not previously been feasible with high-performing models due to deterministic predictions or a lack of conditioning interfaces. Our prompt-based design makes such interventions trivial.

\noindent These interventional insights need not remain descriptive. Just as hypotheses derived from DeepGaze have dramatically improved mechanistic models like SceneWalk~\cite{dagostino2022}, DeepGaze3.5-VL offers a richer substrate for this paradigm, potentially revealing complex behavioral signals to inform future mechanistic architectures.

\paragraph{A Bitter Lesson in Gaze.} This work is an instantiation of \emph{Sutton's Bitter Lesson}~\cite{sutton2019} in the context of modeling human visual attention. Scanpath modeling has long favored domain-specific inductive biases: engineering inhibition-of-return mechanisms, branching spatial and temporal pathways, and hard-coding oculomotor constraints. Our findings suggest visual attention modeling is no exception to Sutton's premise: that general methods leveraging computation and scale consistently overtake human-designed architectures. Rather than relying on custom geometric priors, reframing gaze prediction entirely as autoregressive token generation recovers complex behavioral phenomena purely from data. The strong correlation between multimodal reasoning capability and scanpath predictability reinforces this point. We do not claim that mathematical understanding of visual behavior is pointless, but rather that it should serve as a way to explain model behavior, not as a constraint on model design. Our framework offers a rich, general-purpose substrate for precisely this kind of scientific inquiry.

\paragraph{Toward Single-Pass Density Prediction.}
Our coordinate tokenization requires a prefix-tree traversal with $1{,}111$ passes for exact saliency maps. This is a principled design choice given the available data: each scanpath contains only ${\sim}8$ fixations on average, and many coordinate values (particularly near image edges) appear rarely in natural viewing data. With such sparse coordinate-level statistics, learning dedicated embeddings for each of $10K$ possible $(x,y)$ pairs would be severely underdetermined. However, with sufficiently large coordinate corpora \emph{not necessarily restricted to fixations}, an alternative becomes viable: adding each of the $10K$ coordinate pairs directly to the model's vocabulary as atomic tokens. Under this scheme, predicting the next fixation reduces to a single forward pass producing logits over the full coordinate vocabulary, with log-sum-exp normalization yielding an exact spatial probability map. This would eliminate the prefix-tree computation entirely, reducing full-grid evaluation from $1{,}111$ passes to $1$ pass per fixation, making real-time saliency computation practical.

\section{Conclusion}
We cast scanpath prediction as autoregressive token generation, aligning the task directly with a VLM's native objective and yielding substantial empirical gains. The autoregressive framework absorbs diverse conditioning variables through simple prompt modifications rather than bespoke architectural components. We demonstrate this with three distinct axes: fixation duration modeling, subject identity conditioning, and search target specification, each requiring only changes to the input or predictive text while sharing the same model backbone and training procedure. The analyses in Sections 3.4--3.6 reveal that the model recovers cognition-aligned phenomena from duration statistics alone, captures individual differences in viewing strategy, and redirects spatial predictions when given a search target. These behaviors emerge purely from data, without the explicit oculomotor priors or hand-crafted architectural modules.

\noindent \textbf{Limitations.} Our coordinate discretization to a $100 \times 100$ grid caps spatial resolution at 1\% of image extent; finer grids would require longer token sequences and more prefix-tree passes. This can be addressed as discussed in the previous section. While predictions align with known human visual behavior, the framework does not reveal mechanisms for said behavior without additional analysis.

\noindent \textbf{Broader impact.} Accurate scanpath models are useful in assistive technology, interface design, and clinical screening for neurological conditions. The interventional capability demonstrated in this work offers a tool for generating predictions that would be difficult to elicit experimentally, potentially accelerating oculomotor research while reducing the need for costly human data collection.


\section*{Acknowledgements}
Susmit Agrawal was supported by the German Research Foundation (DFG): SFB 1233, Robust Vision: Inference Principles and Neural Mechanisms, TP C2, project number: 276693517, and thanks the International Max Planck Research School for Intelligent Systems (IMPRS-IS) for support. All authors were supported by the T\"{u}bingen AI Center.

%
%
\bibliographystyle{splncs04}
\bibliography{main}

\clearpage
\newpage
\setcounter{page}{1}
\section{Appendix}
\definecolor{jsonkey}{RGB}{160, 40, 40}
\definecolor{jsonstr}{RGB}{25, 100, 185}
\definecolor{jsonnum}{RGB}{20, 130, 60}
\definecolor{codegray}{RGB}{55, 55, 65}
\definecolor{boxbg}{RGB}{250, 250, 252}
\definecolor{boxborder}{RGB}{180, 185, 210}
\definecolor{titlebg}{RGB}{60, 80, 140}
\definecolor{titlebg1}{RGB}{80, 55, 130}   
\definecolor{titlebg2}{RGB}{50,  110, 70}    
\definecolor{titlebg3}{RGB}{130, 60,  30}    

\lstdefinelanguage{json}{
  basicstyle=\small\ttfamily\color{codegray},
  columns=flexible,
  keepspaces=true,
  breaklines=true,
  showstringspaces=false,
  moredelim=[is][\color{jsonkey}\bfseries]{@k@}{@k@},
  moredelim=[is][\color{jsonstr}]{@s@}{@s@},
  moredelim=[is][\color{jsonnum}]{@n@}{@n@},
}

\subsection{Few-Shot Adaptation to New Datasets}
\label{sec:app_fewshot}

\begin{wrapfigure}{r}{0.45\textwidth}
    \vspace{-2em}
    \centering
    \includegraphics[width=\linewidth]{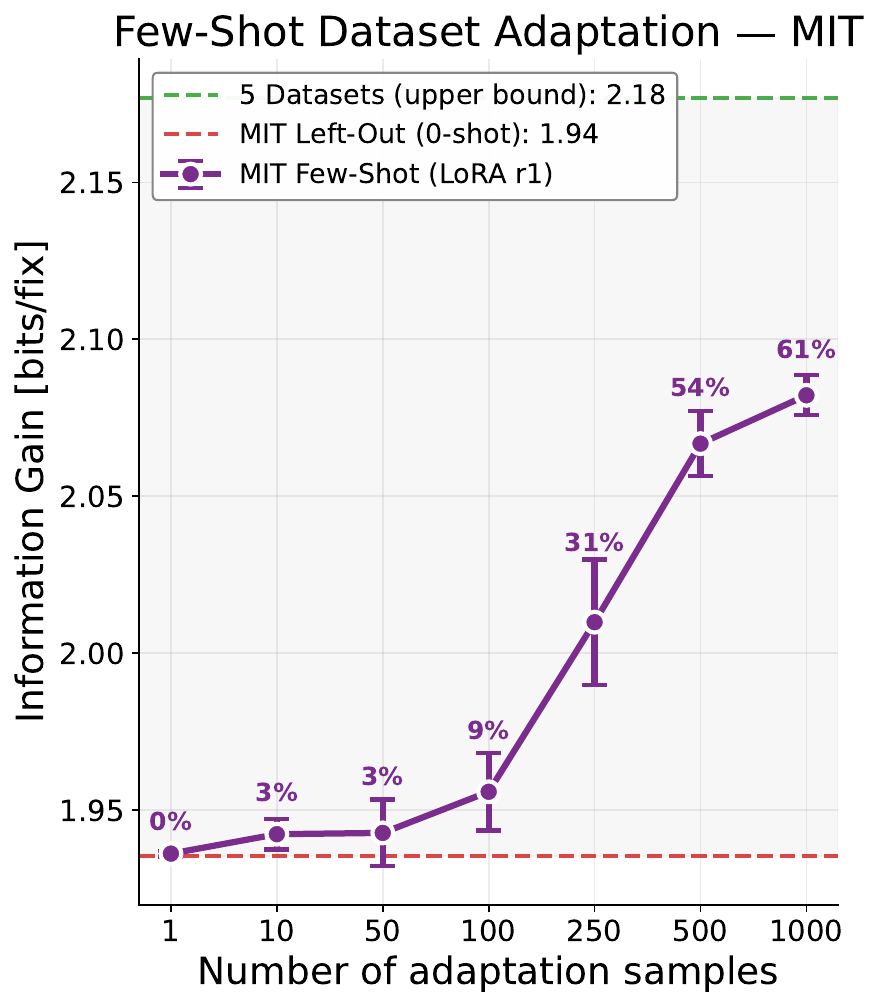}
    \caption{\textbf{Few-shot dataset adaptation.} Fine-tuning a rank-1 LoRA adapter on a held-out dataset (MIT1003) rapidly closes the gap to the fully in-distribution upper bound, confirming that the tuned model requires only minimal calibration to new viewing conditions.}
    \label{fig:fsadapt}
    \vspace{-2em}
\end{wrapfigure}

While out-of-distribution performance is inherently strong (as shown in the LODO benchmarks of the main text), models can be further adapted to the viewing conditions of novel datasets using minimal data. Because the model relies on text-based conditioning rather than hard-coded geometric heads, adapting to a novel dataset's specific viewing biases \cite{khanuja2025modeling} only requires fine-tuning a lightweight rank-1 LoRA adapter rather than retraining an entire architecture. We demonstrate this through few-shot LoRA adaptation on a held-out dataset (Figure~\ref{fig:fsadapt}). Starting from a model trained on the remaining four datasets, we fine-tune using $k$ samples from the held-out dataset's training split, then evaluate on its validation split. Tuning a rank-1 LoRA adapter on a few examples rapidly closes the gap to the fully in-distribution upper bound - about a 1000 scanpath samples (<7\% of the IID training set) closes over 60\% of the gap to the model trained on all 5 datasets combined. Rapid adaptation highlights the flexibility of the autoregressive approach in practical deployment scenarios where large-scale eye-tracking data is unavailable.

\subsection{Extended Subject-Conditioned Analysis}
\label{sec:app_subjectwise}

\begin{figure*}[t]
    \centering
    \includegraphics[width=\linewidth]{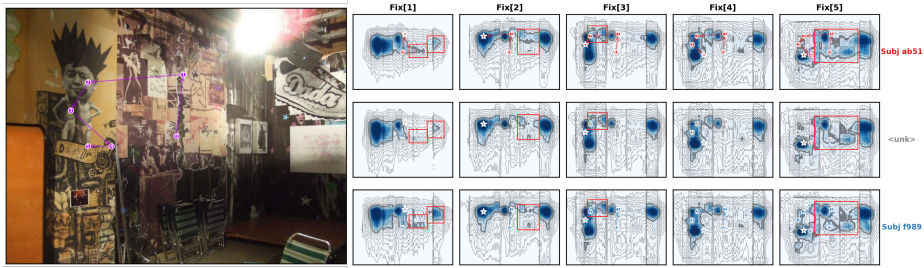}
    \caption{\textbf{Subject-conditioned predictions diverge over successive fixations.} Left: Input image with a scanpath sampled from the subject-agnostic (\texttt{\textless unknown\textgreater}) model. Right: predicted density maps at Fix[1]--[5] for subjects ab51 (top), the \texttt{\textless unknown\textgreater} baseline (middle), and f989 (bottom). All three rows receive identical scanpath context per predicted fixation; differences arise solely from the subject token. Subject-conditioned maps diverge visibly, and the \texttt{\textless unknown\textgreater} prediction hedges across both subjects' preferred regions. Red boxes highlight regions of notable divergence by visual inspection.}
    \label{fig:subject_temporal}
\end{figure*}

As discussed in Section~\ref{sec:subject_conditioning}, conditioning the VLM on subject identity improves predictive accuracy. This improvement stems from the model learning distinct spatial priors and exploratory strategies per observer. Figure~\ref{fig:subject_temporal} details how these subject-specific priors evolve dynamically over the course of a scanpath. When constrained to the exact same past fixation history, predictions conditioned on different subjects remain similar during early orienting fixations but diverge visibly as viewing progresses. For instance, on the image in Figure~\ref{fig:subject_temporal}, while the \texttt{\textless unknown\textgreater} prediction hedges across multiple regions of interest, predictions for subjects ab51 and f989 progressively commit to different viewing strategies. The magnitude of this inter-subject divergence correlates with image ambiguity: on cluttered scenes with multiple competing regions of interest, pairwise Jensen-Shannon Divergence between subject predictions reaches 0.076. On simpler scenes with a single dominant object, divergence remains near zero, indicating the model correctly predicts that structural salience overrides individual idiosyncrasies when an image affords only one obvious viewing strategy.

\subsection{Detailed Visual Search Performance by Target}
\label{sec:app_search}

\begin{figure}[ht]
    \vspace{-1em}
    \centering
    \begin{minipage}{0.48\textwidth}
        \centering
        \includegraphics[width=\linewidth]{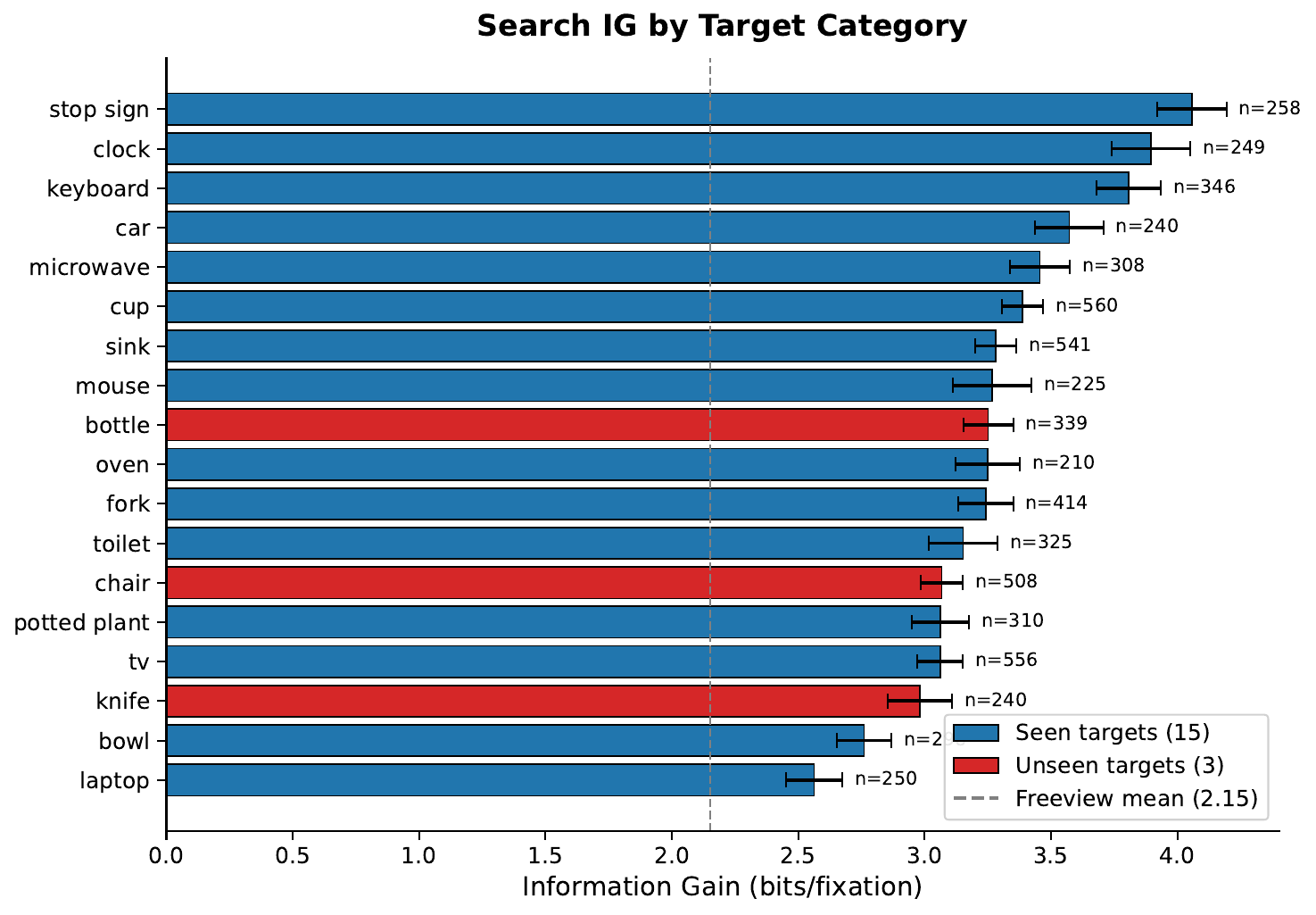}
    \end{minipage}\hfill
    \begin{minipage}{0.48\textwidth}
        \centering
        \includegraphics[width=\linewidth]{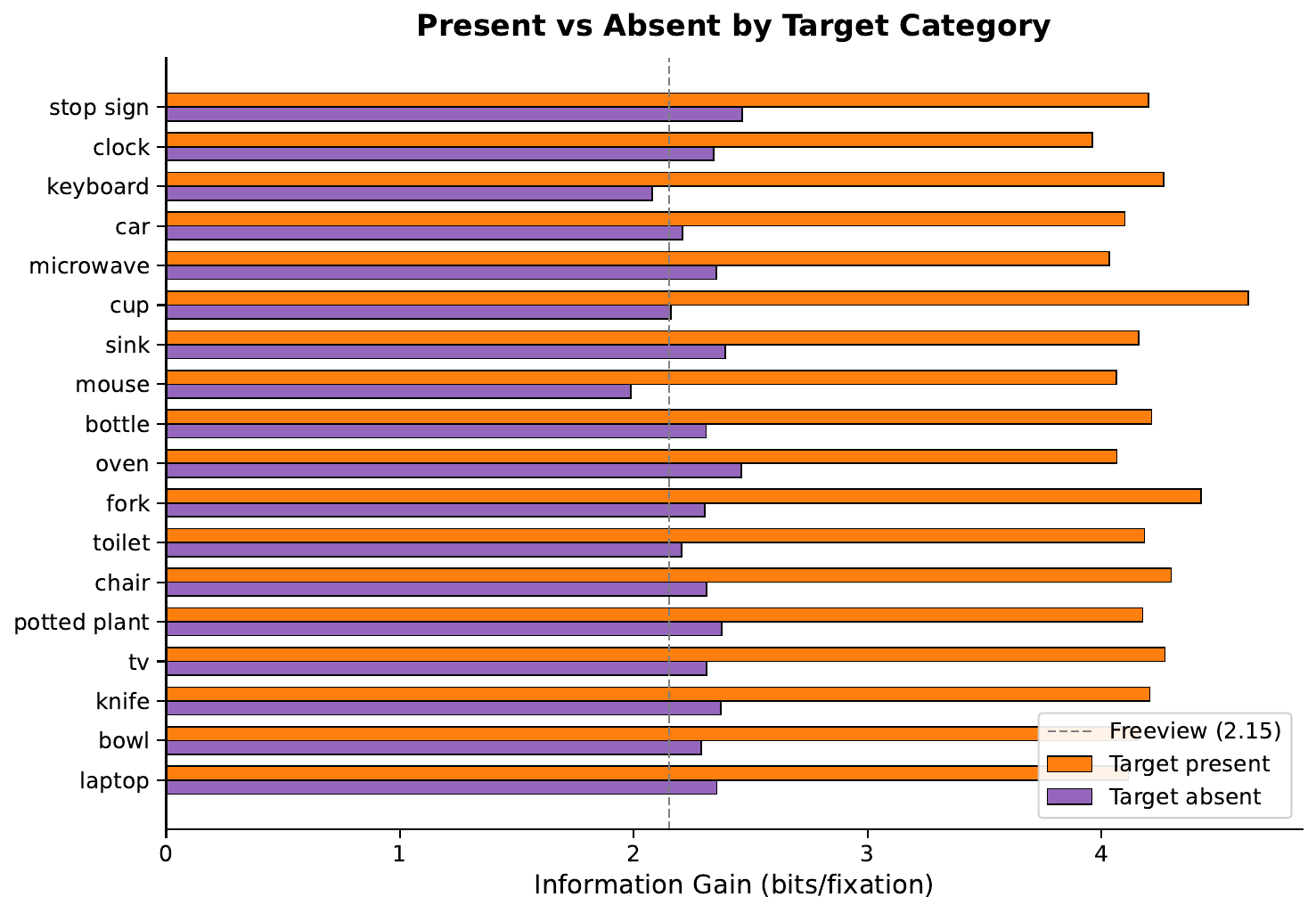}
    \end{minipage}
    \vspace{-10pt}
    \caption{\textbf{Visual Search Performance by Target.} Left: Information gain across categories, including zero-shot generalization to unseen targets. Right: Breakdown of IG when the target is present vs.\ absent in the scene. Unseen target categories (e.g., \textit{bottle, chair, knife}) achieve comparable IG (around 3.0--3.3 bits) to seen targets, demonstrating the VLM's ability to leverage open-vocabulary semantic knowledge.}
    \label{fig:target_ig}
\end{figure}

As described in Section~\ref{sec:search}, the predictive advantage of the visual search model relies heavily on visual guidance when the target is present in the scene. Figure~\ref{fig:target_ig} breaks down Information Gain by target category. In all categories, target-present IG vastly outperforms target-absent IG. This holds true even when the target in question has not been seen during training, and all information about it comes from the model's implicit knowledge base. 

\subsection{Additional Qualitative Results}
\label{sec:app_qualitative}

\begin{figure*}[ht]
    \centering
    \includegraphics[width=\linewidth]{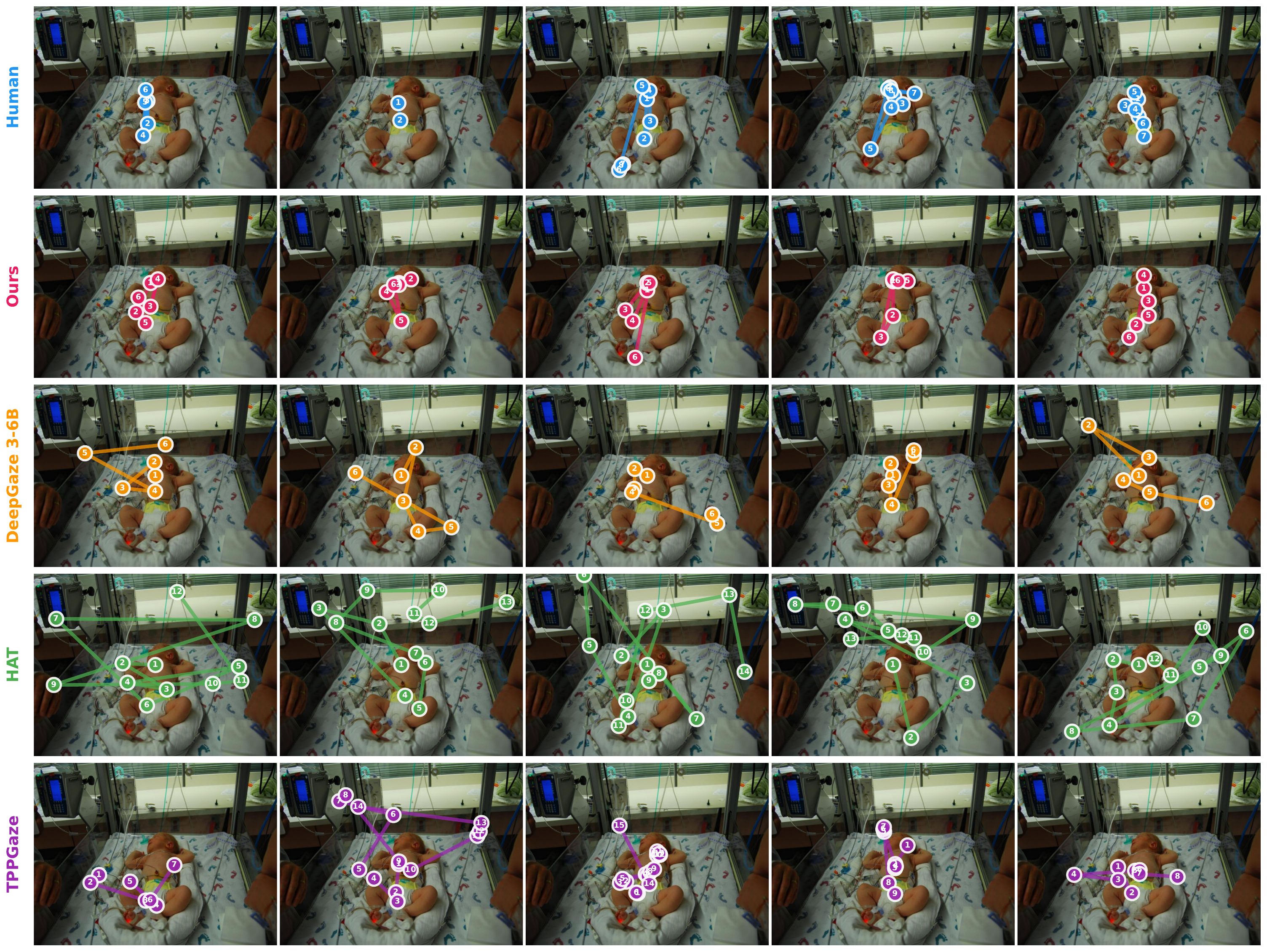}
    \caption{\textbf{Qualitative comparison of generated scanpaths.} Scanpaths sampled from human observers (top row) and various models on an example image from the MIT dataset. DeepGaze3.5-VL (second row) is qualitatively significantly closer to human scanpaths. Both humans and our model consistently scan vertically over the main subject in the image (the baby). In contrast, baseline models (DeepGaze 3-6B, HAT, and TPPGaze in the subsequent rows) often sample peripheral regions, such as the medical equipment and bed edges. While plausible from a purely visual salience perspective, this unconstrained exploratory behavior is not consistent with the targeted, specific viewing behavior exhibited by humans.}
    \label{fig:qualitative_mit0999}
\end{figure*}

Figure~\ref{fig:qualitative_mit0999} provides an extended qualitative comparison of scanpath generation. As observed in the human scanpaths (top row), observers presented with this scene demonstrate high consistency, primarily scanning vertically over the infant. DeepGaze3.5-VL closely matches this behavior, successfully modeling the sustained focus on the primary subject. By comparison, while other models like DeepGaze 3-6B, HAT, and TPPGaze do eventually identify the subject, their generated sequences are characterized by frequent early jumps to peripheral and background regions. While sampling these high-contrast peripheral regions might be visually plausible, it fails to capture the coherent, semantic prioritization consistent with genuine human viewing behavior.

\begin{figure*}[ht]
    \centering
    \includegraphics[width=\linewidth]{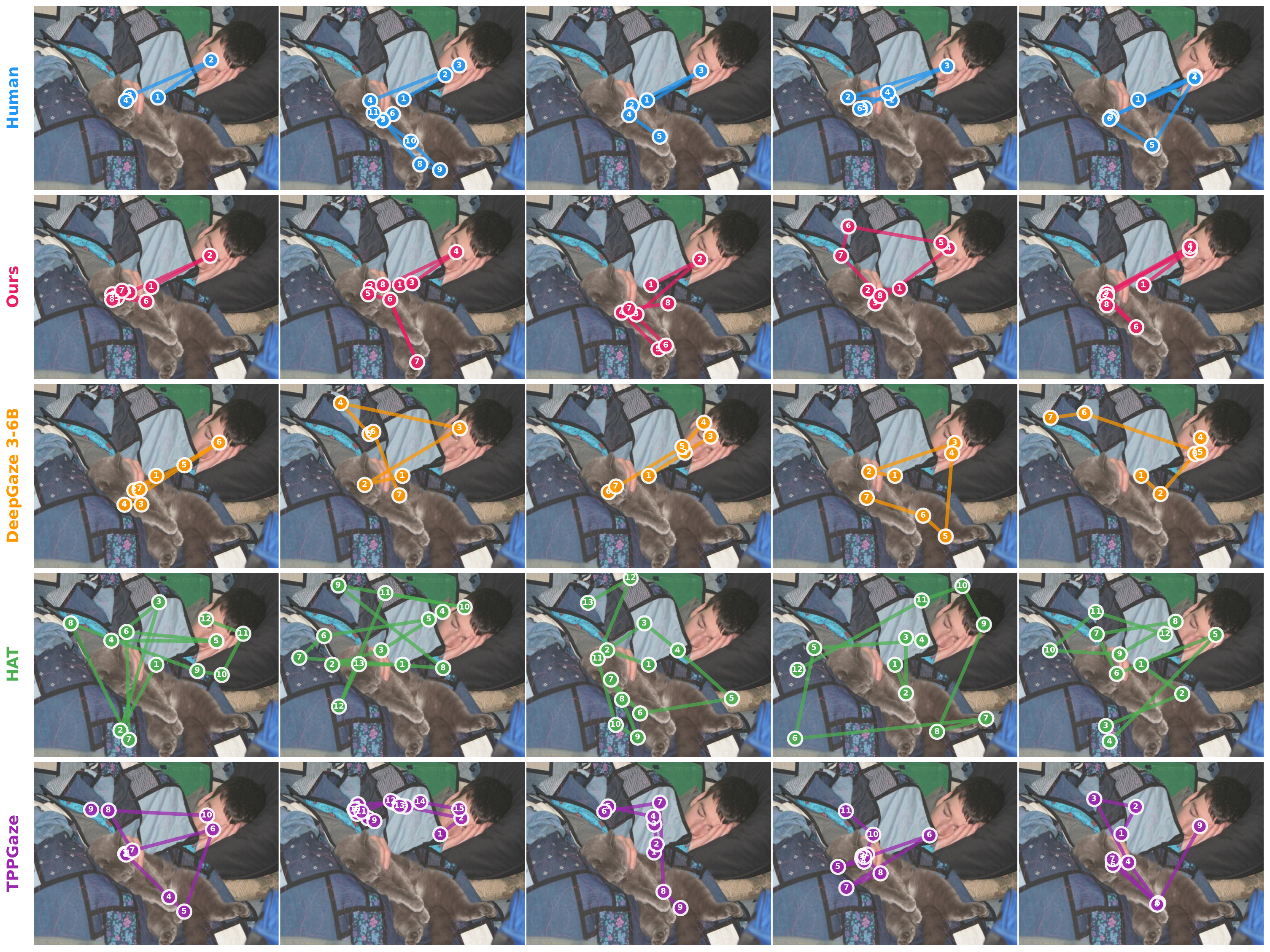}
    \caption{\textbf{Qualitative comparison of distractor suppression.} A more complex scene featuring two semantic subjects against a textured, high-contrast background. Human observers (top row) alternate fixations between the two semantic targets (the person's face and the cat) with few glimpses to the geometric patterns. DeepGaze3.5-VL successfully recovers this semantic behavior. Baseline models fail to suppress the visual distractors, heavily sampling the geometric patterns on the quilt.}
    \label{fig:qualitative_mit0987}
\end{figure*}

Figure~\ref{fig:qualitative_mit0987} further illustrates this capability on a more challenging scene where a strong geometric distractor (the patterned quilt) competes with multiple semantic subjects. Human observers (top row) exhibit a consistent viewing strategy, alternating fixations almost exclusively between the person's face and the cat, with occasional glimpses to the background geometric patterns. DeepGaze3.5-VL's predictions similarly align with this focus on semantic subjects and fleeting glimpses of background patterns. In contrast, baseline models fail to suppress the background texture as effectively.

\subsection{Alignment-Based Scanpath Metrics on COCO-Search18}
\label{sec:app_scanpath_metrics}

In addition to the information-theoretic metrics used throughout the main paper, we report alignment-based scanpath metrics on COCO-Search18~\cite{chen2021coco_search18} for comparison with prior work. Table~\ref{tab:scanpath_metrics_appendix} presents ScanMatch (SM), MultiMatch (MM), String Edit Distance (SED), Fixation Edit Distance (FED), Sequence Score (SS), and Semantic Sequence Score (SemSS) for target-present (TP) and target-absent (TA) conditions, comparing our model against Gazeformer~\cite{mondal2023gazeformer} and GazeXplain~\cite{mondal2023gazexplain}. DeepGaze3.5-VL achieves state-of-the-art performance on the majority of these metrics across both conditions.

\begin{table}[h!]
\centering
\caption{\textbf{Alignment-based scanpath metrics on COCO-Search18.} Results for target-present (TP) and target-absent (TA) conditions. SED uses the grid-based computation following GazeXplain; FED uses the cluster-based approach from Gazeformer. Best results are in \textbf{bold}.}
\label{tab:scanpath_metrics_appendix}
\scriptsize
\begin{tabular}{lcccccc}
\toprule
Model & SM($\uparrow$) & MM($\uparrow$) & SED($\downarrow$) & FED($\downarrow$) & SS($\uparrow$) & SemSS($\uparrow$) \\
\midrule
Gazeformer~\cite{mondal2023gazeformer}-TP & 0.43 & 0.80 & 2.22 & 2.06 & 0.47 & 0.39 \\
GazeXplain~\cite{mondal2023gazexplain}-TP & 0.48 & 0.81 & \textbf{1.98} & 2.10 & 0.54 & 0.44 \\
DeepGaze3.5-VL-TP & \textbf{0.60} & \textbf{0.89} & 2.21 & \textbf{1.67} & \textbf{0.65} & \textbf{0.47} \\
\midrule
Gazeformer~\cite{mondal2023gazeformer}-TA & 0.35 & 0.81 & 4.49 & 3.46 & 0.36 & 0.35 \\
GazeXplain~\cite{mondal2023gazexplain}-TA & 0.37 & 0.81 & 4.31 & 4.22 & 0.38 & 0.36 \\
DeepGaze3.5-VL-TA & \textbf{0.44} & \textbf{0.85} & \textbf{3.83} & \textbf{2.57} & \textbf{0.58} & \textbf{0.40} \\
\bottomrule
\end{tabular}
\end{table}

\paragraph{Caveats on alignment-based metrics.} While we report these metrics for completeness, we note several important limitations that motivate our preference for information-theoretic evaluation throughout the main paper.

First, alignment-based metrics evaluate \emph{point-predicted} sequences rather than the inherent uncertainty in viewing behaviour that scanpath models should express (which can only be measured probabilistically). This promotes regression to a ``mean scanpath''~\cite{kummerer2022state,kummerer2015information} and penalizes models that faithfully represent the stochastic variability of human gaze.

Second, these metrics depend on arbitrary parameter choices in their computation. A striking example is the discrepancy between SED and FED in Table~\ref{tab:scanpath_metrics_appendix}. Both measure scanpath similarity by converting fixation sequences to strings and computing edit distances, differing only in how they assign string alphabets (grid-based for SED vs.\ cluster-based for FED). Despite this seemingly minor implementation difference, they yield markedly divergent results: DeepGaze3.5-VL achieves the best FED by a wide margin while GazeXplain leads on SED. The point-prediction problem affects both metrics equally, yet the arbitrary alphabet assignment produces contradictory model rankings.

Information Gain, by contrast, provides principled, model-agnostic comparability~\cite{kummerer2015information}. As a linear measure of information, IG retains discriminative power even in regimes where bounded ranking metrics like AUC saturate near their ceiling. Any conclusions drawn from alignment-based metrics should therefore be interpreted with caution.

\subsection{Implementation and Training Details}
\label{sec:app_implementation}

\paragraph{Architecture and LoRA configuration.} 
We employ parameter-efficient fine-tuning through Low-Rank Adaptation (LoRA)~\cite{lora} as our primary training method. LoRA freezes pretrained weights and introduces trainable low-rank decomposition matrices $W' = W + \Delta W = W + BA$, where $W \in \mathbb{R}^{d \times k}$ is the frozen pretrained weight matrix, and $B \in \mathbb{R}^{d \times r}$, $A \in \mathbb{R}^{r \times k}$ are trainable matrices with rank $r \ll \min(d, k)$. This reduces trainable parameters from $d \times k$ to $r \times (d + k)$ while preserving the model's pretrained knowledge.

We apply Low-Rank Adaptation (LoRA) to all MLP layers in the base LLM. For single-dataset experiments, we use rank $r=8$ with an alpha scaling factor $\alpha = 16$ and dropout of $0.05$. For experiments jointly training across multiple datasets, we increase capacity to $r=32$ with $\alpha = 64$. 

\paragraph{Optimization schema.} 
All models are trained with AdamW optimization ($\beta_1 = 0.9$, $\beta_2 = 0.999$). We employ a cosine learning rate decay schedule with a linear warmup period of 10\% of the first epoch. The model trains for a maximum of 10 epochs with a per-device batch size of 2 and gradient accumulation to achieve an effective batch size of 16. Validation cross-entropy is monitored after every epoch, and we use early stopping when validation loss hits a minimum to prevent overfitting. These hyperparameters are kept fixed across all experiments.

\paragraph{Datasets.} We use 5 different datasets for robust evaluation of our framework.
\begin{itemize}
    \item \textbf{MIT1003} It comprises 1,003 mostly photographic indoor and outdoor images, viewed by 15 observers for 3 seconds each. We pool all scanpaths that have more than 1 fixation, and use this pool to create our train/validation splits. We ensure that the splits are at an image level, not scanpath level - no images in the validation set are seen during training. Our split yields 14,775 training scanpath samples and 270 evaluation samples.

    \item \textbf{CAT2000:}
    It contains 2,000 images uniformly distributed across diverse semantic categories, with fixations recorded from 24 observers over 5 seconds. We use a similar image-level split as MIT, yielding 33,470 training scanpaths and 678 validation scanpaths.

    \item \textbf{COCO-FreeView:} It comprises 4317 high-resolution natural scenes, viewed by 10 observers for 5 seconds each. Our training/validation splits contain 42,188 and 860 scanpaths, respectively.

    \item \textbf{DAEMONS:} It comprises 2,200 high-resolution natural scenes, viewed by 10 to 50 observers (sampled from a total pool of 250 participants) for 8 seconds each. Our training/validation splits contain 27,792 and 2917 scanpaths, respectively.

    \item \textbf{Figrim:} It comprises 2,787 natural scene images across 21 indoor and outdoor categories, viewed for 2 seconds each by 11 to 22 observers, depending on the image type. Our training/validation splits contain 42,864 and 867 scanpaths, respectively.
\end{itemize}

All baselines are evaluated on the same validation sets, and DeepGaze3-6B uses the same training splits as described above. We provide sample prompts used for training DeepGaze3.5-VL in a Free-Viewing setting in Figure~\ref{fig:sample-free}, in a Subject-Conditioned Setting in Figure~\ref{fig:training-example}, in a joint location and duration modeling setting in Figure~\ref{fig:sample-durations}, and for visual search in Figure~\ref{fig:sample-search}.

\paragraph{Inference.} We use the vLLM framework for accelerated inference,  taking under a minute for IG computation using the rapid evaluation strategy described in Section~\ref{sec:methods} on our MIT validation set (270 scanpaths). Sampling a scanpath from the model - the primary inference use-case - takes the same time as the rapid evaluation method. For AUC computation, inference with vLLM takes roughly 2 hours on a single GPU for prefix-tree evaluation (see Figure~\ref{fig:schematic} for a visualization of both procedures).

\begin{figure}[ht]
\centering
\begin{tcolorbox}[
  enhanced, arc=3pt, boxrule=1pt,
  colback=boxbg, colframe=boxborder,
  title={\sffamily\small\bfseries\color{white}
         Sample Data for Free-Viewing Scanpath Prediction},
  colbacktitle=titlebg1,
  left=10pt, right=10pt, top=6pt, bottom=8pt,
]
\begin{lstlisting}[language=json]
{
  @k@"conversations"@k@: [
    {
      @k@"from"@k@: @s@"human"@s@,
      @k@"value"@k@: @s@"<image>Analyze this image and predict a human eye movement scanpath during free viewing for 3 seconds.@s@
@s@    A scanpath is the temporal sequence of fixation points showing where a person looks over time. Consider visual saliency, semantic importance, and how attention naturally flows across a scene.@s@

@s@    Generate a scanpath of exactly 10 fixation points in temporal order as a list of tuples: (x, y)@s@
@s@        - x: horizontal position (0-100, 0=left, 100=right)@s@
@s@        - y: vertical position (0-100, 0=top, 100=bottom)@s@
@s@        - Points should be ordered from first to last fixation.@s@

@s@    Output ONLY a Python list of tuples: [(51,46),(38,28),...]"@s@
    },
    {
      @k@"from"@k@: @s@"gpt"@s@,
      @k@"value"@k@: @s@"[(50, 48), (21, 48), (09, 46), (35, 48), (54, 49),@s@
@s@                (82, 48), (92, 50), (95, 56), (59, 48), (35, 45)]"@s@
    }
  ],
  @k@"images"@k@: [@s@"/path/to/image.jpg"@s@]
}
\end{lstlisting}
\end{tcolorbox}
\caption{Free-viewing scanpath prediction. The model predicts $(x,y)$ fixation coordinates
only, without duration. JSON keys in \textcolor{jsonkey}{\textbf{red}}, string values in \textcolor{jsonstr}{blue}.}
\label{fig:sample-free}
\end{figure}

\begin{figure}[ht]
\centering

\begin{tcolorbox}[
  enhanced, arc=3pt, boxrule=1pt,
  colback=boxbg, colframe=boxborder,
  title={\sffamily\small\bfseries\color{white} Sample Data for Free-Viewing Scanpath Prediction with Subject Conditioning},
  colbacktitle=titlebg,
  left=10pt, right=10pt, top=6pt, bottom=8pt,
]
\begin{lstlisting}[language=json]
{
  @k@"conversations"@k@: [
    {
      @k@"from"@k@: @s@"human"@s@,
      @k@"value"@k@: @s@"<image>Analyze this image and predict a human eye movement scanpath during free viewing for 3 seconds by subject 33d20163.@s@  
@s@    A scanpath is the temporal sequence of fixation points showing where a person looks over time. Consider visual saliency, semantic importance, and how attention naturally flows across a scene.@s@

@s@    Generate a scanpath of exactly 8 fixation points in temporal order as a list of tuples: (x, y)@s@
@s@        - x: horizontal position (0-100, 0=left, 100=right)@s@
@s@        - y: vertical position (0-100, 0=top, 100=bottom)@s@
@s@        - Points should be ordered from first fixation to last fixation.@s@

@s@    Output ONLY a Python list of tuples:\n[(51,46),(38,28),...]"@s@
    },
    {
      @k@"from"@k@: @s@"gpt"@s@,
      @k@"value"@k@: @s@"[(52, 48), (47, 56), (22, 48), (13, 45),@s@
@s@                (56, 43), (78, 51), (79, 55), (56, 44)]"@s@
    }
  ],
  @k@"images"@k@: [@s@"/path/to/image.jpg"@s@]
}
\end{lstlisting}
\end{tcolorbox}

\caption{A single training instance for scanpath prediction fine-tuning. JSON keys in
\textcolor{jsonkey}{\textbf{red}}, string values in \textcolor{jsonstr}{blue}.}
\label{fig:training-example}
\end{figure}

\vspace{12pt}

\begin{figure}[ht]
\centering
\begin{tcolorbox}[
  enhanced, arc=3pt, boxrule=1pt,
  colback=boxbg, colframe=boxborder,
  title={\sffamily\small\bfseries\color{white}
         Sample Data for Free-Viewing Scanpath Prediction with Fixation Durations},
  colbacktitle=titlebg2,
  left=10pt, right=10pt, top=6pt, bottom=8pt,
]
\begin{lstlisting}[language=json]
{
  @k@"conversations"@k@: [
    {
      @k@"from"@k@: @s@"human"@s@,
      @k@"value"@k@: @s@"<image>Analyze this image and predict a human eye movement scanpath during free viewing for 3 seconds.@s@
@s@    A scanpath is the temporal sequence of fixation points showing where a person looks over time. Consider visual saliency, semantic importance, and how attention naturally flows across a scene.@s@

@s@    Generate a scanpath of exactly 10 fixation points in temporal order, with fixation duration, as a list of tuples: (x, y, t)@s@
@s@        - x: horizontal position (0-100, 0=left, 100=right)\n@s@
@s@        - y: vertical position (0-100, 0=top, 100=bottom)\n@s@
@s@        - t: fixation duration in milliseconds (0-999)\n@s@
@s@        - Points should be ordered from first to last fixation.@s@

@s@    Output ONLY a Python list of tuples:\n[(51,46,221),(38,28,754),...]"@s@
    },
    {
      @k@"from"@k@: @s@"gpt"@s@,
      @k@"value"@k@: @s@"[(50, 48, 521), (21, 48, 142), (09, 46, 100), (35, 48, 183), (54, 49, 250), (82, 48, 096), (92, 50, 233), (95, 56, 162), (59, 48, 321), (35, 45, 075)]"@s@
    }
  ],
  @k@"images"@k@: [@s@"/path/to/image.jpg"@s@]
}
\end{lstlisting}
\end{tcolorbox}
\caption{Free-viewing scanpath prediction with fixation durations. The model predicts
$(x,y,t)$ tuples, where $t$ is dwell time in milliseconds.}
\label{fig:sample-durations}
\end{figure}

\vspace{12pt}

\begin{figure}[ht]
\centering
\begin{tcolorbox}[
  enhanced, arc=3pt, boxrule=1pt,
  colback=boxbg, colframe=boxborder,
  title={\sffamily\small\bfseries\color{white}
         Sample Data for Target-Directed Visual Search Scanpath Prediction},
  colbacktitle=titlebg3,
  left=10pt, right=10pt, top=6pt, bottom=8pt,
]
\begin{lstlisting}[language=json]
{
  @k@"conversations"@k@: [
    {
      @k@"from"@k@: @s@"human"@s@,
      @k@"value"@k@: @s@"<image>Analyze this image and predict a human eye movement scanpath while searching for a clock.@s@
@s@    A scanpath is the temporal sequence of fixation points showing where a person looks over time. Consider the search target, visual saliency, and how attention naturally flows during visual search.@s@

@s@    Generate a scanpath of exactly 2 fixation points in temporal order as a list of tuples: (x, y)\n@s@
@s@        - x: horizontal position (0-100, 0=left, 100=right)\n@s@
@s@        - y: vertical position (0-100, 0=top, 100=bottom)\n@s@
@s@        - Points should be ordered from first to last fixation.@s@

@s@     Output ONLY a Python list of tuples:\n[(51,46),(38,28),...]"@s@
    },
    {
      @k@"from"@k@: @s@"gpt"@s@,
      @k@"value"@k@: @s@"[(50, 50), (50, 24)]"@s@
    }
  ],
  @k@"images"@k@: [@s@"/path/to/image.jpg"@s@]
}
\end{lstlisting}
\end{tcolorbox}
\caption{Target-directed visual search scanpath prediction. The search target
(here: \textit{clock}) is specified in the prompt; the number of fixations varies
per trial reflecting actual search termination.}
\label{fig:sample-search}
\end{figure}



\end{document}